\documentclass[lettersize,journal]{IEEEtran}
\usepackage{amsmath,amsfonts}

\usepackage{algorithm}
\usepackage{algpseudocode}

\usepackage{array}
\usepackage[caption=false,font=normalsize,labelfont=sf,textfont=sf]{subfig}
\usepackage{textcomp}
\usepackage{stfloats}
\usepackage{url}
\usepackage{verbatim}
\usepackage{graphicx}
\usepackage{cite}
\usepackage{url}
\usepackage{tabularx}
\usepackage{booktabs}
\usepackage{color}
\usepackage[table]{xcolor} 
\usepackage{multirow}

\begin{document}

\title{SIDeR: Semantic Identity Decoupling for Unrestricted Face Privacy}

\author{Zhuosen Bao, Xia Du, Zheng Lin, Jizhe Zhou, Zihan Fang, Jiening Wu, Yuxin Zhang, Zhe Chen, Chi-man Pun,~\IEEEmembership{Senior Member,~IEEE}, Wei Ni,~\IEEEmembership{Fellow,~IEEE}, and Jun Luo,~\IEEEmembership{Fellow,~IEEE}
\thanks{Zhuosen Bao and Xia Du are with the School of Computer and Information Engineering, Xiamen University of Technology, Xiamen, 361000, China (email: baozhuosen@stu.xmut.edu.cn; duxia@xmut.edu.cn).}
\thanks{Zheng Lin is with the Department of Electrical and Electronic Engineering, University of Hong Kong, Pok Fu Lam, Hong Kong SAR, China (e-mail: linzheng@eee.hku.hk).}
\thanks{Jizhe Zhou is with the School of Computer Science, Engineering Research Center of Machine Learning and Industry Intelligence, Sichuan University, Chengdu, China, 610020, China (email: jzzhou@scu.edu.cn).}
\thanks{Zihan Fang is with the Department of Computer Science, City University
of Hong Kong, Kowloon, Hong Kong SAR, China (e-mail: zihanfang3-c@my.cityu.edu.hk).}
\thanks{Jiening Wu is with the College of Artificial Intelligence, Southwest University, Chongqing 400715, China (e-mail: jnwu66@swu.edu.cn).}
\thanks{Yuxin Zhang and Zhe Chen are
with the School of Computer Science, Fudan University, Shanghai
200438, China (e-mail: ail: yxzhang24@m.fudan.edu.cn; zhechen@fudan.edu.cn). }
\thanks{Chi-man Pun is with the Department of Computer and Information Science, Faculty of Science and Technology, University of Macau, Macau, 999078, China (email: cmpun@umac.mo).}
\thanks{W. Ni is with Data61, CSIRO, Marsfield, NSW 2122, Australia, and the School of Computing Science and Engineering, and the University of New South Wales, Kennington, NSW 2052, Australia (e-mail:
wei.ni@ieee.org).}
\thanks{J. Luo is with the College of Computing and Data Science, Nanyang Technological University, Singapore (e-mail: junluo@ntu.edu.sg).}
}

\markboth{Journal of \LaTeX\ Class Files,~Vol.~14, No.~8, August~2021}%
{Shell \MakeLowercase{\textit{et al.}}: A Sample Article Using IEEEtran.cls for IEEE Journals}


\maketitle

\begin{abstract}
With the deep integration of facial recognition into online banking, identity verification, and other networked services, achieving effective decoupling of identity information from visual representations during image storage and transmission has become a critical challenge for privacy protection. To address this issue, we propose SIDeR, a Semantic decoupling–driven framework for unrestricted face privacy protection. SIDeR decomposes a facial image into a machine-recognizable identity feature vector and a visually perceptible semantic appearance component. By leveraging semantic-guided recomposition in the latent space of a diffusion model, it generates visually anonymous adversarial faces while maintaining machine-level identity consistency. The framework incorporates momentum-driven unrestricted perturbation optimization and a semantic–visual balancing factor to synthesize multiple visually diverse, highly natural adversarial samples. Furthermore, for authorized access, the protected image can be restored to its original form when the correct password is provided. Extensive experiments on the CelebA-HQ and FFHQ datasets demonstrate that SIDeR achieves a 99\% attack success rate in black-box scenarios and outperforms baseline methods by 41.28\% in PSNR-based restoration quality. 
\end{abstract}

\begin{IEEEkeywords}
Face privacy protection, semantic decoupling, diffusion models, adversarial attack.
\end{IEEEkeywords}

\section{Introduction}

\IEEEPARstart{T}{he} rapid evolution of deep neural networks (DNNs)~\cite{sun2025rrto,duan2025leed,yuan2025constructing,zhao2024leo,lin2025sl,fang2025dynamic,zhang2025robust}, particularly through Generative Adversarial Networks (GANs)\cite{NIPS2014_f033ed80,lin2022channel} and Diffusion Models\cite{NEURIPS2020_4c5bcfec,11245609}, has fundamentally transformed generative modeling by enabling the synthesis of hyper-realistic visual content that is often indistinguishable from authentic data. However, these advancements represent a double-edged sword; while they empower creative industries, they simultaneously escalate security and privacy concerns through the rise of sophisticated deepfake technologies. By exploiting facial data—a uniquely sensitive biometric modality—malicious actors can now fabricate lifelike forged media to impersonate individuals, compromise authentication systems, and erode digital trust. Consequently, protecting facial biometrics against synthetic exploitation has emerged as a critical imperative in the modern information security landscape.

\begin{figure}[t]
    \centering
    \includegraphics[width=1\linewidth]{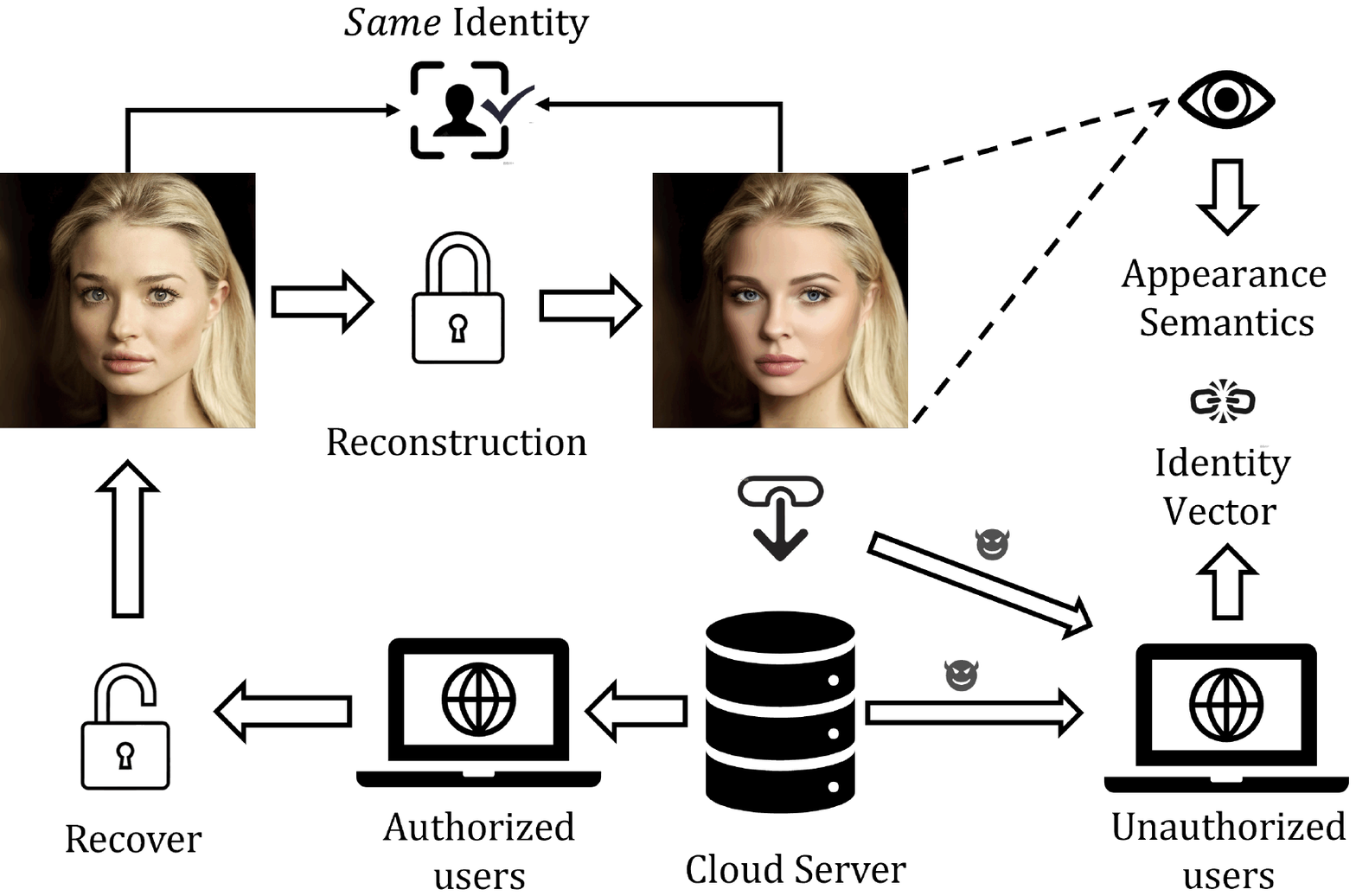}
    \caption{Illustration of \textit{\textbf{visual-machine divergence}} where the machine-level identity is preserved through adversarial anchoring. The protected image maintains machine-level identity consistency with the original while supporting authorized reversible recovery. Unauthorized access disrupts the alignment between visual semantics and identity features.}
    \label{fig:motivation}
\end{figure}

In this context, we consider a specific yet insufficiently defined problem, referred to in this paper as unrestricted face privacy protection. Unlike conventional privacy-preserving approaches that enforce explicit visual similarity constraints, predefined perturbation budgets, or irreversible identity removal, unrestricted face privacy enables visually unconstrained modification of facial appearance while preserving machine-level identity consistency and supporting authorized recovery when required. Under this setting, protected faces may appear visually unrelated to their original appearance to human observers, yet remain consistently verifiable by trusted face recognition systems.

A substantial body of research has focused on face de-identification, which suppresses identifiable characteristics through image-level obfuscation or generative reconstruction \cite{gross2006model,meden2017kappa,chen2007tools,butler2015privacy,gu2020password,li2023riddle,hukkelaas2019deepprivacy,maximov2020ciagan}. Although these approaches can effectively reduce human-observable identity cues, they often disrupt the identity representations required by automated face recognition systems, making it difficult to simultaneously ensure privacy protection and maintain machine-level identity consistency.
More recently, adversarial perturbation–based approaches have been explored for facial privacy protection by exploiting the vulnerability of face recognition models \cite{hu2022protecting,shamshad2023clip2protect,li2024transferable,shamshad2024makeup,sun2024diffam}. These methods aim to prevent unauthorized machine-based identification while preserving perceptual appearance. However, most existing adversarial approaches are primarily designed for attack-oriented objectives, such as inducing misclassification or verification failure, and provide limited support for identity-consistent protection or reversible restoration.

Despite their effectiveness in reducing recognition accuracy, adversarial attack–based approaches overlook a fundamental requirement of practical biometric systems: legitimate identity verification must remain functional. In many real-world scenarios, the face recognition system itself is considered a trusted authentication module that must reliably verify the identities of authorized users. Consequently, privacy protection mechanisms should not undermine the system’s ability to recognize the genuine user when authentication is intended.

This leads to an inherent contradiction: on one hand, strong visual obfuscation is often necessary to suppress privacy leakage from human observers; on the other hand, such obfuscation tends to destroy the identity information required by machine-based verification. Conversely, preserving discriminative identity cues for recognition typically reveals sufficient visual detail to enable unauthorized inspection. This intrinsic conflict between visual privacy and machine-level identity consistency remains a central challenge in deployable face privacy protection, yet existing research provides limited solutions that reconcile both objectives under realistic constraints. Most prior methods implicitly assume mutual exclusion between privacy protection and identity verification, preventing the design of mechanisms that support both privacy-preserving views and identity-consistent machine representations.
In response to this challenge, emerging visual identity–hiding paradigms explore whether identity information can be re-encoded at the visual level while preserving access to machine-recognizable identity representations. The goal is to generate facial renderings that appear natural yet visually anonymous to humans, while remaining verifiable by trusted recognition systems. An overview of this paradigm is illustrated in Fig.~\ref{fig:motivation}.

In this paper, we conceptualize facial privacy protection as the challenge of deconstructing and controllably reconstructing identity information. Within this paradigm, a facial image is decomposed into two functionally independent components: a machine-recognizable identity representation and human-perceptible visual semantics. Such disentanglement enables fine-grained control over the identity content preserved in the reconstructed output. By selectively reconfiguring the visual semantic component while retaining or transforming the identity representation, the system can generate fully anonymized faces for privacy-preserving data sharing or produce protected faces that remain verifiable exclusively by authorized recognition models for secure authentication. The key advantage of this formulation lies in its controllability, which enables seamless adjustment between anonymity and verifiability to meet different privacy–utility requirements, thereby providing a principled foundation for deployable, adaptive face privacy protection.

To address the aforementioned challenges, we introduce SIDeR (Semantic–Identity Decoupled Unrestricted Face Privacy Protection), a unified generative framework for reversible and identity-consistent facial privacy preservation. SIDeR leverages a semantic identity decomposition paradigm in which the digital identity is anchored within the adversarial manifold of face recognition networks, while the visual semantics are substantially reconfigured through diffusion-based generative priors. This design ensures that the protected face is visually anonymous to humans yet remains machine-recognizable to authorized authentication systems. By explicitly decoupling identity features from human-visible facial appearance, SIDeR enables the synthesis of photorealistic, identity-protected faces that achieve a principled balance between privacy and recognition utility.
To further enhance robustness against unauthorized identification, SIDeR incorporates a momentum-driven unrestricted perturbation strategy together with a semantic–visual balancing mechanism. These components synthesize visually diverse, natural-looking adversarial variants while maintaining consistent machine-level identity representations. 

Specifically, we first introduce a semantic guidance mechanism that precisely identifies identity-related facial regions, ensuring that privacy perturbations are applied to semantically critical attributes rather than indiscriminately modifying the entire image. We further develop a momentum-guided latent perturbation strategy that directly optimizes adversarial objectives within the latent representation space. This allows SIDeR to generate identity-consistent yet semantically diverse facial renderings under text-driven control, thereby achieving explicit decoupling between identity features and human-visible semantics.
To support identity restoration in authorized scenarios, inspired by reversible neural networks~\cite{guan2022deepmih}, we design a reversible embedding mechanism which incorporates the structural alignment transformation of the original face into the anonymization process. This enables near-lossless recovery of the source face when a valid authorization key is provided during storage or transmission, while maintaining strong visual anonymity for unauthorized access. Together, these components allow SIDeR to provide flexible, reversible, and privacy-preserving identity protection without undermining machine-level recognition fidelity.

In summary, our main contributions are as follows:
\begin{itemize}
\item \textbf{Semantic–identity decoupling for controllable anonymization.}
We propose a novel semantic identity decoupling mechanism that disentangles machine-recognizable identity features from human-perceptible appearance semantics in the latent space, thereby protecting user privacy while maintaining high utility for machine-driven identity verification. Together with a momentum-driven unrestricted perturbation strategy, SIDeR generates photorealistic, visually anonymous faces while preserving high image fidelity.
\item \textbf{Latent-space generative privacy with reversible identity restoration.}
We introduce a unified latent-space privacy generation paradigm that jointly enables semantic disentanglement and conditional reversibility. By performing adversarial optimization directly in the latent manifold, SIDeR produces semantically consistent anonymized faces while establishing an efficient bijective transformation that supports authorized, near-lossless restoration of the original identity.
\item \textbf{State-of-the-art privacy robustness, visual fidelity, and recovery accuracy.}
Extensive experiments across multiple datasets and face recognition systems demonstrate that SIDeR consistently outperforms existing methods, achieving up to 99\% attack success rate in black-box settings and improving PSNR-based recovery quality by over 41\% compared with representative baselines. These results indicate that SIDeR provides a deployable and identity-consistent solution for face privacy protection.
\end{itemize}

The remainder of this paper is organized as follows. Section II reviews related work on face privacy protection. Section III introduces the proposed SIDeR framework and its key components. Section IV presents experimental evaluations. Section V discusses ablation studies and additional analyses. Section VI concludes the paper.


\section{Related Work}
Adversarial attacks have emerged as an important research direction in the field of deep learning security, revealing the vulnerability of neural network models within the input space\cite{goodfellow2015explainingharnessingadversarialexamples}. An adversarial attack refers to the addition of carefully crafted and subtle perturbations to an input sample. These perturbations are almost imperceptible to the human eye yet are sufficient to cause a model to produce incorrect predictions or identity classifications\cite{11267085, zhu2024dp}. Such perturbations are typically obtained via optimization procedures that aim to maximize prediction error or classification loss while constraining the magnitude of the perturbation so that the image's perceptual quality remains unchanged.

In recent years, the concept of adversarial attacks has been adopted for facial privacy protection. This line of research leverages adversarial perturbations to preserve the natural and realistic appearance of a face image while preventing automated recognition systems from correctly identifying the individual. Early approaches \cite {goodfellow2014explaining,dong2018boosting,madry2017towards,dong2019evading,yang2021towards} focused on adding constrained noise perturbations directly in the pixel domain to mislead face recognition models. However, these methods often introduce visible artifacts that undermine visual naturalness. Subsequent studies explored unrestricted attacks that relax the magnitude constraints imposed on perturbations to enhance transferability and concealment. Representative approaches include methods that embed perturbations into cosmetic patterns applied to specific facial regions~\cite{komkov2021advhat, yin2021adv, hu2022protecting, shamshad2023clip2protect, li2024transferable, shamshad2024makeup, sun2024diffam}, and methods that employ generative models to synthesize adversarial examples that remain visually consistent with the original face~\cite{liu2023diffprotect, zhang2024double, an2024sd4privacy, liu2024adv, chow2025personalized, wang2025adv}. Although these approaches aim to suppress machine recognition while preserving human-perceivable facial content, they primarily target recognition models and become ineffective in scenarios involving facial forgery or generative reconstruction, where visual features dominate the synthesis process. Moreover, most of these techniques lack reversibility, making it impossible to recover the original image in legitimate use cases.

In contrast, the idea of visual identity hiding has gained increasing attention. This research direction seeks to conceal visual identity features to the extent that human observers cannot perceive them, while maintaining automated systems' ability to perform recognition. Inspired by adversarial attack principles, Su \textit{et al.}~\cite{su2023hiding} proposed a method for hiding visual identity information by overlaying a noise layer on the original image through an interaction between a recognition and a recovery model, enabling both encryption and restoration. Wang \textit{et al.}~\cite{Wang_Liu_Luo_Yang_Wang_2022} decomposed facial images into frequency components and selected channels that are relevant to identity recognition yet irrelevant to visual reconstruction. They further introduced a fast masking strategy to protect residual frequency content. 

Although such methods successfully obscure visual identity, their noise-based modifications can still be detected by human observers. To alleviate this issue, Zhang \textit{et al.}~\cite{10121472} integrated a stream cipher-based attribute confusion module with an attribute adversarial network to achieve a reversible facial attribute privacy protection mechanism. This mechanism enables the concealment of multiple soft-biometric attributes under user-controlled keys while preserving identity verifiability and supporting secure restoration of the original attributes. They \textit{et al.}~\cite{he2024diff} employed pretrained diffusion models to generate visually altered facial images, thereby reducing training overhead. Nevertheless, existing methods either neglect the need for reversible recovery or degrade the quality of the restored images compared with the originals.

\begin{figure*}[htbp]
    \centering
    \begin{center}
    \includegraphics[width=1\linewidth]{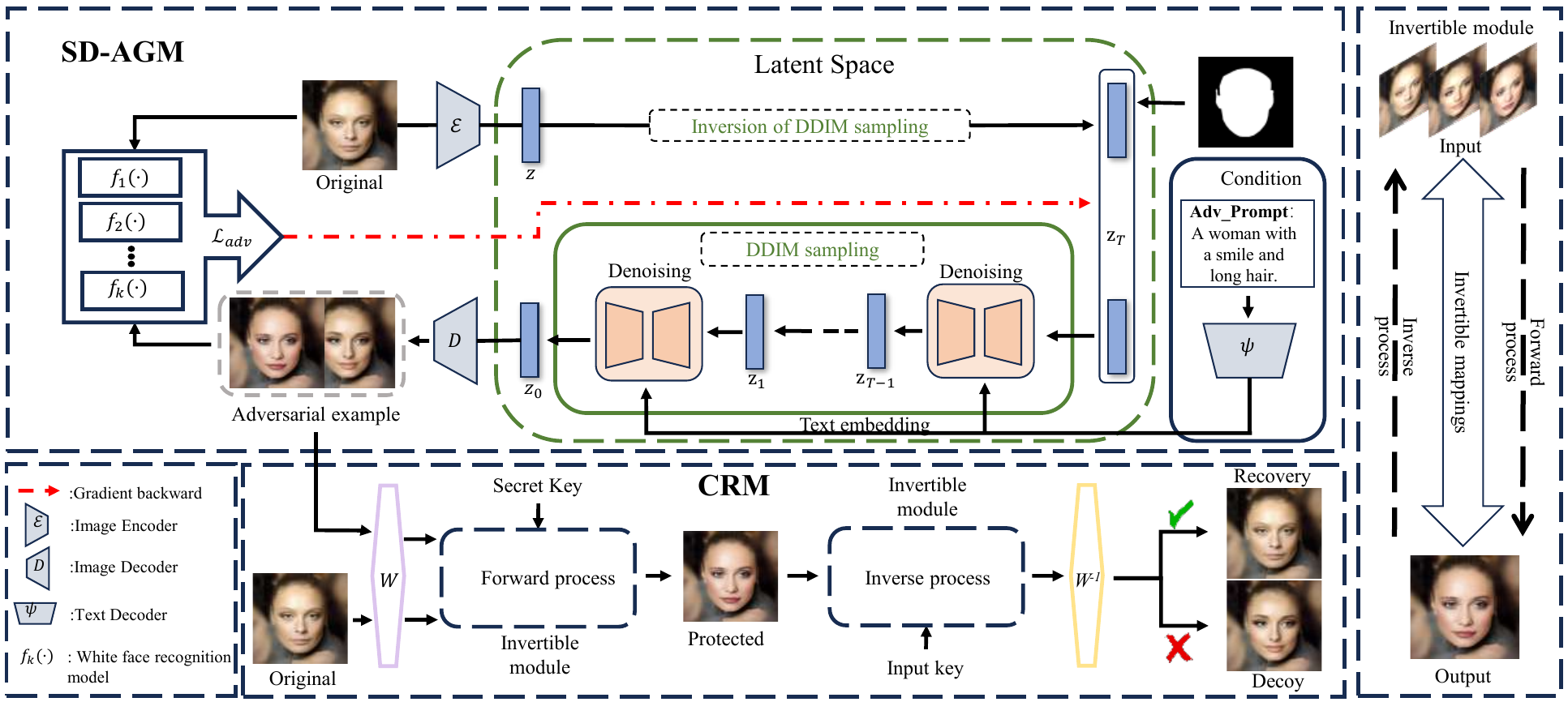}
    \end{center}
    \caption{Achieve visual identity information hiding through deconstruction of facial identity, with the protected image maintaining identity consistency with the original at the machine recognition level, while supporting reversible recovery under legitimate authorization; if maliciously stolen by unauthorized users, the consistency of appearance semantics and visual identity will be lost.}
    \label{fig:framework}
\end{figure*}

\section{Proposed SIDeR Framework}
This section presents the proposed SIDeR framework, a unified generative paradigm for reversible and semantically controllable facial privacy protection. The design of SIDeR is motivated by a fundamental observation underlying unrestricted face privacy protection: human perception and machine-based face recognition rely on inherently different representations of facial identity. While humans interpret identity through semantic visual attributes, modern face recognition systems encode identity in a learned embedding space largely invariant to perceptual variations. Based on this observation, SIDeR adopts semantic–identity decoupling as its core principle, explicitly separating machine-recognizable identity representations from human-perceptible semantic appearance. This allows visually unconstrained yet semantically coherent identity anonymization while maintaining identity consistency for authorized recognition, and avoids treating privacy protection as a mere trade-off between visual distortion and recognition accuracy.

Built upon this principle, the SIDeR framework consists of two major components: the \textbf{Semantic-Decoupled Adversarial Generation Module (SD-AGM)} and the \textbf{Conditionally Reversible Module (CRM)}. SD-AGM manipulates latent representations within a diffusion model under semantic guidance to generate visually distinct yet identity-preserving facial images, while CRM embeds the original facial information into the anonymized carrier through an invertible mechanism, enabling authorized, high-fidelity recovery without exposing identity cues to unauthorized users. An overview of the pipeline is illustrated in Fig.~\ref{fig:framework}.

\subsection{Problem Definition}\label{define}
We consider the problem of unrestricted and reversible face privacy protection. Let $x\in\mathbb{R}^{H\times W\times3}$ denote an input facial image containing both identity-discriminative information and semantic appearance attributes. The goal of SIDeR is to synthesize a privacy-preserving face $\hat{x}$ that exhibits perceptible visual deviation from the original face to human observers while maintaining identity consistency under face recognition models and allowing exact reconstruction of the original image through an authorized inverse mapping.

A high-quality protected sample must therefore satisfy three conditions.
First, the synthesized image  $\hat{x}$ should be visually distinct from $x$ according to human perception, expressed as $H(\hat{x})\neq H(x)$, where $H(\cdot)$ denotes human perceptual judgment. Second, despite the perceptual alteration, $\hat{x}$ must preserve the identity representation under a face recognition model $F:\mathcal{X}\times\mathcal{X}\to\mathbb{R}$, where larger values of $\mathcal{F(\cdot)}$ indicate greater identity similarity. Formally,
\begin{equation}\label{eq1}
\mathcal{F}(\hat{x},x)\ge\tau,
\end{equation}
with verification threshold $\tau$. Finally, SIDeR incorporates a key-conditioned invertible embedding module $I_k(\cdot, \cdot)$. Let $x_{gen}$ denote the visually anonymized face synthesized by the generative module $G$. The final protected image $\hat{x}$ is generated by embedding the original image $x$ into $x_{gen}$ as a hidden signal:
\begin{equation}\label{eq2}
\hat{x} = I_k(x_{gen}, x),
\end{equation}
where $x_{gen}$ serves as the visual cover. The inverse mapping $I_k^{-1}(\cdot)$ enables the exact recovery of the original identity from $\hat{x}$ using the correct key:
\begin{equation}\label{eq3}
x = \mathcal{R}\left(I_k^{-1}(\hat{x})\right),
\end{equation}
where $\mathcal{R}(\cdot)$ extracts the hidden information.
By jointly satisfying perceptual discrepancy, machine-level identity preservation, and strict key-conditioned bidirectional reversibility, SIDeR supports controlled and user-authorized identity reveal while preventing unauthorized visual inspection.

\subsection{Preliminaries}
\textbf{Stable Diffusion.} Stable Diffusion (SD)~\cite{rombach2022high} is a latent diffusion model that performs image synthesis by iteratively denoising a latent variable within a learned low-dimensional representation space. Following\cite{kingma2013auto}, a variational autoencoder (VAE) is used to map images into this latent space. Let \(x\) denote an input image and let \(z_0=\mathcal{E}(x)\) be its latent representation produced by the VAE encoder \(\mathcal{E}\). Starting from this latent, the forward diffusion progressively corrupts \(z_0\) into \(z_1,\dots,z_T\) according to the fixed Markov transitions:
\begin{equation}\label{eq4}
    q(z_{t}\mid z_{t-1})={\mathcal{N}}\Big({\sqrt{1-\beta_{t}}}\,z_{t-1},\,\beta_{t}I\Big)\,,\qquad t=1,\ldots,T,
\end{equation}
where $\beta_{t}$ is a predefined variance schedule controlling the noise magnitude at each step. 
This process gradually transforms the latent variable into an approximate isotropic Gaussian distribution.

The reverse process aims to invert the corruption by predicting the noise that has been added. It is parameterized by a neural network $\epsilon_\theta$, yielding the following transition distribution:
\begin{equation}\label{eq5}
p_\theta(z_{t-1}\mid z_t,c)=\mathcal{N}(\mu_\theta(z_t,t,c),\mathrm{~}\Sigma_\theta(z_t,t,c)),
\end{equation}
where the conditioning input $c$ incorporates semantic information such as text embeddings or other structural priors. During training, the network learns to approximate the true score function that removes noise from different diffusion levels.
To accelerate sampling, Stable Diffusion adopts the deterministic DDIM sampling scheme, which bypasses stochasticity while preserving the consistency of the diffusion trajectory. The latent update at step 
$t \to t - 1$is given by
\begin{equation}\label{eq6}
z_{t-1}=\sqrt{\alpha_{t-1}}\hat{x}_0(z_t,t,c)+\sqrt{1-\alpha_{t-1}}\epsilon_\theta(z_t,t,c),
\end{equation}
with 
\begin{equation}\label{eq7}
\alpha_t=\prod_{i=1}^t(1-\beta_i),\quad\hat{x}_0(z_t,t,c)=\frac{z_t-\sqrt{1-\alpha_t}\epsilon_\theta(z_t,t,c)}{\sqrt{\alpha_t}},
\end{equation}
here, $\hat{x}_0$ denotes the model’s estimate of the clean latent before noise injection.

SD employs classifier-free guidance to strengthen the influence of conditioning information. By combining the  conditional $c$ and unconditional $\emptyset$ noise predictions, the guided score is expressed as
\begin{equation}\label{eq8}
\epsilon_{\mathrm{guided}}=\epsilon_\theta(z_t,t,c)+s\left(\epsilon_\theta(z_t,t,c)-\epsilon_\theta(z_t,t,\emptyset)\right),
\end{equation}
where $s$ is the guidance scale that controls the trade-off between semantic alignment and sample diversity. Larger values of $s$ amplify the conditional signal and typically produce more faithful but less diverse outputs. Finally, the potential variable $\mathbf{z}_{0}$ after de-noising is mapped back to the pixel space by the VAE decoder $\mathcal{D}(\cdot)$.

All diffusion and sampling operations are performed in the latent space learned by the VAE, which significantly reduces computational cost while retaining sufficient semantic capacity. This latent-space modeling strategy enables Stable Diffusion to generate high-resolution, photorealistic images with efficient inference.

\textbf{Identity–Semantic Decomposition.}
SIDeR relies on a disentangled representation of facial information that separates identity-discriminative cues from human-perceptible appearance semantics. Unlike traditional latent disentanglement that relies on structural constraints, SIDER achieves functional decoupling by optimizing two divergent objectives: 1) Visual Semantic Shift, which drives the pixel-level appearance away from the source; and 2) Identity Anchoring, which forces the adversarial features to remain within the source's manifold in the recognition space. Given an input face image $x$, the identity embedding is extracted as $z_{\mathrm{id}}=\mathcal{F}(x)$. In parallel, a semantic appearance descriptor is obtained from the SD’s latent encoder $z_{\mathrm{sem}}=E(x)$, where $E(\cdot)$ maps the image to a structured latent space capturing appearance attributes such as texture, expression, hairstyle, and illumination.

This decomposition yields two complementary latent factors $z_{id}$, which preserve machine-level identity consistency, and $z_{sem}$, which supports controllable visual modification without altering identity. Treating identity and appearance as disentangled components enables SIDeR to perform semantically guided generation and adversarial optimization within the diffusion latent space, forming the basis for unrestricted yet perceptually coherent facial anonymization.
\subsection{Semantic-Decoupled Adversarial Generation Module}
Based on the latent variable at diffusion step $t$, the deterministic DDIM update in Equations \eqref{eq6}–\eqref{eq7} defines a single denoising transition from $z_t$ to $z_{t-1}$.
We formalize the entire iterative reverse process as a composite mapping $\Omega(\cdot)$ that applies these transitions sequentially over $T$ steps.
\begin{equation}\label{eq:omega}
\Omega(z_T, T, c, \emptyset)
= \Phi_1\big( \Phi_2( \cdots \Phi_T(z_T, T, c, \emptyset ) ) \cdots \big),
\end{equation}
where each transformation $\Phi_t$ corresponds to a single DDIM denoising update:
\begin{equation}\label{eq:phi}
\begin{split}
\Phi_t(z_t, t, c, \emptyset)
&= \sqrt{\alpha_{t-1}}\hat{x}_0(z_t,t,c,\emptyset) \\
&\quad + \sqrt{1-\alpha_{t-1}}\epsilon_{\mathrm{guided}}(z_t,t,c,\emptyset).
\end{split}
\end{equation}
Since the VAE decoder is fully differentiable, the diffusion trajectory represented by $\Omega(\cdot)$ fully determines the synthesis result in latent space.

The efficacy of the adversarial generation hinges on a high-fidelity conditioning prompt $P$ that guides the synthesis of photorealistic face images. To ensure the prompt is visually descriptive and structurally optimized, we implement a two-stage automated prompt generation strategy. First, we utilize Image-to-Text models to automatically generate an initial descriptive caption $\hat{P}$ corresponding to the input image $x$. Subsequently, $\hat{P}$ is fed into a Large Language Model (LLM), leveraging the latter's advanced editing capabilities to refine and stylize the description into the final, optimized prompt $P$. To interface with the latent diffusion model, the optimized prompt $P$ is transformed into the conditioning embedding $c$ via the text encoder $\psi$, i.e., $c = \psi(P)$. This ensures that the adversarial process employs a semantically rich, structurally consistent conditioning embedding.

We leverage the differentiability of $\Omega(\cdot)$ to embed the adversarial optimization process directly into the initialization of the latent code $z_T$. The objective is to find an optimal initial latent code $z_T^*$ such that the resulting generated adversarial image $x_{adv}=\mathcal{D}(\Omega(z_T^*))$ maximizes the retention of the original image $x'$s
identity features at the machine-recognition level, while adhering to the specified text semantics $c$.

\textbf{Ensemble Identity Loss Function.} To ensure the adversarial sample $x_{adv}$ possesses strong transferability and robustness across various facial recognition systems, we employ an Ensemble Loss $\mathcal{L}_{\mathrm{ens}}$ based on cosine similarity. This function aims to minimize the feature-space distance between $x_{adv}$ and the original image $x$ across a set of heterogeneous facial recognition models $\mathcal{F(\cdot)}$:
\begin{equation}
\mathcal{L}_{\mathrm{adv}}(\mathcal{F}; z_T, x) = \sum_{f \in \mathcal{F}} w_f \cdot \left( 1 - \cos(f(x_{gen}), f(x)) \right),
\end{equation}
here, $f(\cdot)$ denotes the feature vector extracted by the recognition model, and $w_f$ is the weight assigned to each model. Our optimization objective is thus defined as:
\begin{equation}
    \min_{z_T} \mathcal{L}_{\mathrm{ens}}(\mathcal{F}; \mathcal{D}(\Omega(z_T, T, c, \emptyset)); x).
\end{equation}

\textbf{Momentum Latent Space Optimization.} Directly optimizing $\mathcal{L}_{\mathrm{ens}}$ across the entire latent space often leads to gradient oscillation and unstable adversarial perturbations that leak into the background. To enhance the stability and control the perturbation region, we introduce both Momentum and a Facial Masking Constraint.

Let $\mathcal{M} \in \{0, 1\}$ be the binary mask derived from a face parsing module and downsampled to match the dimensionality of the latent code $z_T$. This mask restricts the application of the adversarial gradient solely to the facial region.

In the iterative optimization step $k$, we apply the mask $\mathcal{M}$ to the accumulated momentum $g_{k+1}$ before updating the latent code $z_T^{(k)}$:
\begin{equation}
\begin{aligned}
g_{k+1} &= \mu \cdot g_k + \frac{\nabla_{z_T} \mathcal{L}_{\mathrm{ens}}(z_T^{(k)})}{\|\nabla_{z_T} \mathcal{L}_{\mathrm{ens}}(z_T^{(k)})\|_1}, \\
z_T^{(k+1)} &= z_T^{(k)} - \alpha \cdot \text{sign}(\mathcal{M} \odot g_{k+1}),
\end{aligned}
\end{equation}
where $\mu$ is the momentum decay factor, $g_k$ is the accumulated momentum, and $\alpha$ is the step size.  Here, $\nabla_{z_T}$ denotes the gradient operator with respect to $z_T$, $\text{sign}(\cdot)$ represents the sign function, and $\odot$ denotes the Hadamard product (element-wise multiplication).  This Momentum-based Iterative Gradient Descent stabilizes the process, ensuring that the identity features are reliably guided into $z_T$.

\begin{algorithm}[h]
\caption{SD-AGM}
    \label{alg:sd-agm}
    \begin{algorithmic}[1]
        \Require Input $x$, Prompt $P$, DDIM steps $T$, White FR models $\mathcal{F}_k$, Mask $\mathcal{M}$, Attack iterations $N$, Learning rate $\alpha$, and Momentum factor $\mu$.
        \Ensure Adversarial Pair: $x_{\text{cover}}, x_{\text{decoy}}$.

        \State $z_0 \gets \mathcal{E}(x)$ ; $c \gets \psi(P)$ 

        \For{$j$ = 1 to 2}
            \State $z_T \gets q(z_T \mid z_0)$; $g_0 \gets 0$; $\delta_0 \gets 0$
            \For{$k=1$ to $N$}
                \State $z_{adv} \gets \Omega(z_T^{k-1} , T, c, \emptyset)$
                \State $x_{adv} \gets \mathcal{D}(z_{\text{adv}})$
                \State $g_k \gets \mu \cdot g_{k-1} + \frac{\nabla_{z_T} \mathcal{L}_{\mathrm{ens}}(\mathcal{F};x_{adv},x)}{\|\nabla_{z_T} \mathcal{L}_{\mathrm{ens}}(\mathcal{F};x_{adv},x)\|_1}$
                \State $z_T^k \gets z_T + \alpha \cdot \text{sign}(\mathcal{M} \odot g_k)$ 
            \EndFor
            \State $x^{(j)} \gets x_{adv}$
        \EndFor
        \State \textbf{return} $x_{\text{cover}} \gets x^{(1)}, x_{\text{decoy}} \gets x^{(2)}$
        
    \end{algorithmic}
\end{algorithm}

\textbf{Perceptual Guidance Tuning.} During the optimization process, the strength of Classifier-Free Guidance (CFG), controlled by the scale $s$ in Equation \eqref{eq8}, critically influences the visual quality and adversarial capacity of the generated image. As noted, a large value of $s$ enforces stricter semantic alignment, but this often leads to excessively stringent constraints in the latent space, resulting in visual distortion and inhibiting the smoothness required for successful adversarial perturbation.

To address this, we introduce the Perceptual Guidance Tuning mechanism to precisely regulate the effective CFG intensity, prioritizing visual naturalness and fidelity over absolute semantic rigidity. We introduce a modulation factor $\lambda \in (0, 1]$ to control the contribution of the conditional guidance.

We define the modulated guided score 
$\epsilon_{\mathrm{mod}}$ by applying $\lambda$ to the guidance differential in Equation \eqref{eq8}:
\begin{equation}
\epsilon_{\mathrm{mod}} = \epsilon_\theta(z_t, t, \emptyset) + \lambda \cdot s \cdot (\epsilon_\theta(z_t, t, c) - \epsilon_\theta(z_t, t, \emptyset)).
\end{equation}
By substituting the modulated score $\epsilon_{\mathrm{mod}}$ into $\Phi_t$ and experimentally setting the combined effective guidance scale $\lambda \cdot s$ to a lower range, we achieve the following benefits:1) Lowering $\lambda \cdot s$ softens the generative manifold, creating a larger search space for the identity gradient $\nabla_{z_T} \mathcal{L}_{\mathrm{ens}}$; 2) It avoids visual degradation and artifacts resulting from overemphasis on the text semantics under high guidance, thus significantly improving the perceptual realism of $x_{adv}$.

\textbf{Dual-Sample Generation via Latent Stochasticity.} To construct the Adversarial Decoy required by the subsequent CRM module, SD-AGM must generate a pair of identity-consistent but visually diverse samples $\{x_{cover}, x_{decoy}\}$ for a single input identity $x$.

We utilize the same text prompt $P$ and the same optimization objective $\mathcal{L}_{\mathrm{ens}}$, but we initialize the optimization and generation process with two independent random latent codes $z_{T}^{(1)}$ and $z_{T}^{(2)}$:
\begin{equation}
\begin{aligned}
x_{cover} &= \mathcal{D}(\Omega(z_{T}^{(1)}; x, P, \emptyset)), \\
x_{decoy} &= \mathcal{D}(\Omega(z_{T}^{(2)}; x, P, \emptyset)).
\end{aligned}   
\end{equation}

This approach, rooted in latent space stochasticity, ensures that $x_{cover}$ and $x_{decoy}$ reside on the same semantic manifold, share the same identity features, yet exhibit sufficient visual differentiation. This foundational step is critical for enabling the Nested Identity Masquerading in the CRM. The specific attack algorithm is described in Algorithm \ref{alg:sd-agm}.

\begin{figure*}[htbp]
    \centering
    \includegraphics[width=1\linewidth]{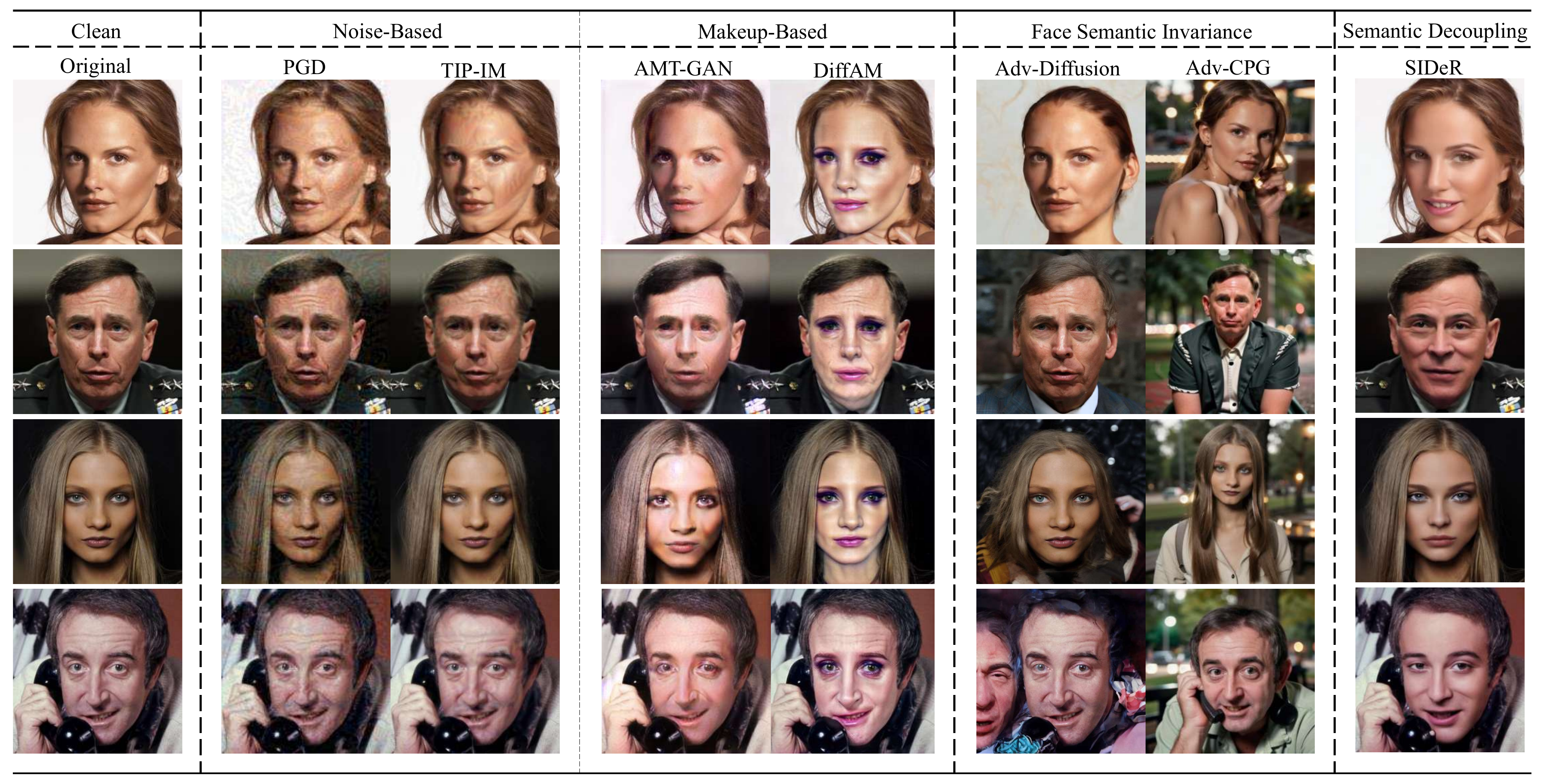}
    \caption{Qualitative comparison of adversarial examples on the CelebA-HQ dataset. Compared to three state-of-the-art adversarial methods, SIDeR generates examples with higher visual fidelity, particularly in facial details, background consistency, and overall naturalness.}
    \label{fig:attack}
\end{figure*}

\begin{figure*}[htbp]
    \centering
    \includegraphics[width=1\linewidth]{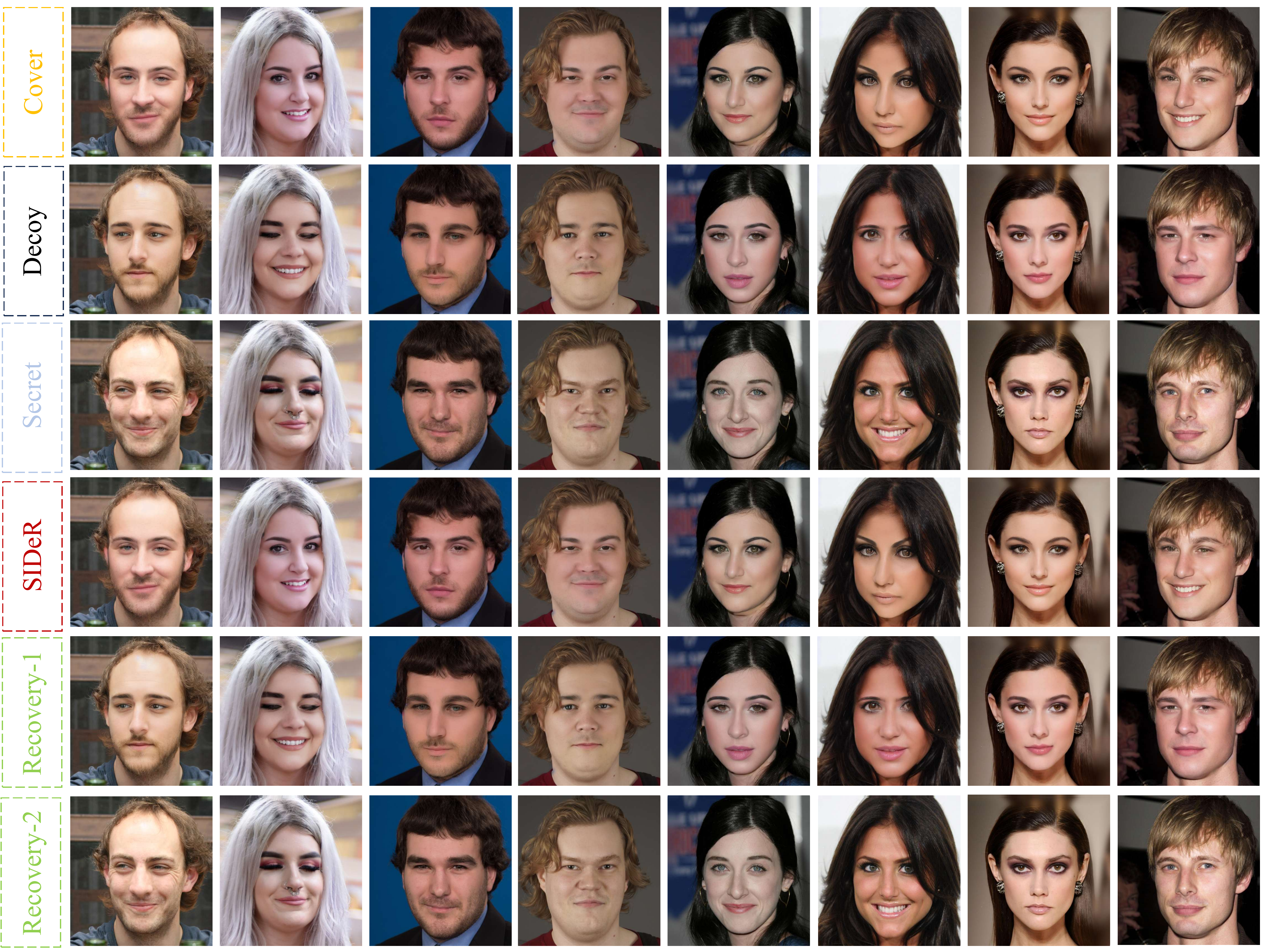}
    \caption{A visualization of SIDeR's multi-level recovery capabilities. The figure demonstrates a high degree of visual consistency between different stages: the top three rows represent the original carrier image "Cover," the decoy image "Decoy," and the secret image to be hidden "Secret," respectively; the "SIDeR" row corresponds to the final generated protected image; "Recover-1" and "Recover-2" represent the extracted intermediate decoy layer and the final recovered secret image, respectively. The outputs of all stages maintain natural image lighting and identity features, with no obvious visual degradation observed.}
    \label{fig:recover_self}
\end{figure*}

\subsection{Conditionally Reversible Module}
The Conditionally Reversible Module (CRM) integrates the dual samples $\{x_{cover}, x_{decoy}\}$ generated by the SD-AGM with the original image $x$, enabling key-gated, deceptive, and near-lossless recovery. Inspired by Invertible Neural Networks (INNs), we adopt a nested bijective architecture to ensure reconstruction fidelity. All images are first transformed into the wavelet domain using the Discrete Wavelet Transform (DWT, $\mathcal{W}$) to enhance hiding capacity and balance computational efficiency.

The CRM implements the abstract embedding function $I_k(\cdot, \cdot)$ through two cascaded networks, $I_{\text{deep}}$ and $I_{\text{shallow}}$, constructing a "First-In, Deepest-Hidden" nested structure.

\textbf{Forward Nested Embedding.} The process begins by embedding the true identity information into the core layer. The first step hides the original private image $x$ into the carrier $x_{cover}$ using the key-gated deep network:
\begin{equation}\label{eq:deep_fwd}
\mathcal{W}(h_{\text{inter}}) = I_{\text{deep}}(\mathcal{W}(x_{cover}), \mathcal{W}(x), k),
\end{equation}
where $\mathcal{W}(h_{\text{inter}})$ is the intermediate latent feature containing the encrypted private identity. Subsequently, the second step embeds the adversarial decoy $x_{decoy}$ into this intermediate feature to create a deceptive outer layer:
\begin{equation}\label{eq:shallow_fwd}
\mathcal{W}(\hat{h}) = I_{\text{shallow}}(\mathcal{W}(h_{\text{inter}}), \mathcal{W}(x_{decoy})).
\end{equation}

The final protected image $\hat{x}$ is obtained via the Inverse Discrete Wavelet Transform (IDWT, $\mathcal{W}^{-1}$): $\hat{x} = \mathcal{W}^{-1}(\mathcal{W}(\hat{h}))$. This cascaded structure ensures that $x$ is wrapped by the decoy layer, making $x_{decoy}$ the first observable signal during the inverse process.

\textbf{Conditional Inverse Mapping.} The core innovation of CRM lies in its $I_{k}^{-1}(\cdot)$ function, which enforces two distinct recovery paths based on the provided key $k$.

\textbf{Unauthorized Access.} When the provided key $k$ is incorrect or absent, the system executes the deception protocol. Following the LIFO (Last-In-First-Out) principle of INNs, the system only performs the first-level (shallow) inverse mapping:
\begin{equation}\label{eq:deception_final} x_{\text{out}} = \mathcal{W}^{-1}\left(I_{\text{shallow}}^{-1}(\mathcal{W}(\hat{x})),  k_{wrong}\right) \to x_{decoy},
\end{equation}
in this path, the attacker recovers the Adversarial Decoy $x_{decoy}$. Unlike traditional steganography which might yield noise, $x_{decoy}$ provides: 1) Visual Discrepancy, misleading human observers into believing the decryption is complete; and 2) Identity Functionality, where the decoy still passes machine recognition ($\mathcal{F}(x_{decoy}, x) \geq \tau$). This achieves an effective honeypot defense.

\textbf{True Recovery.} With the correct key $k$, the system sequentially decapsulates the nested layers to retrieve the secret image $x$:
\begin{equation}\label{eq:authorized_final}
\mathcal{W}(x) = I_{\text{deep}}^{-1}\left( I_{\text{shallow}}^{-1}(\mathcal{W}(\hat{x})), k \right).
\end{equation}
The final output $x_{\text{out}} = \mathcal{W}^{-1}(\mathcal{W}(x))$ achieves the near-lossless reconstruction of the private identity.

\section{Experiment}
\subsection{Experimental Setting}
\textbf{Implementation Details.} The learning rate $\alpha$ is set to 0.01, and all experiments are conducted on an NVIDIA A40 GPU. We employ a 20 steps DDIM sampler with a denoising strength of 0.75, a classifier-free guidance scale $s$ of 1, a momentum factor $\mu$ of 0.6, and a modulation coefficient $\lambda$ of 3. During the attack stage, the guidance prompt can be freely specified by the generator. For efficient and consistent batch processing, we first obtain an initial textual description of each image using BLIP-2\cite{li2023blip}, and subsequently refine it into a concise guidance prompt using GPT-4\cite{achiam2023gpt}.

\textbf{Datasets.} We evaluate SIDeR on two high-quality face benchmarks, CelebA-HQ~\cite{CelebAMask-HQ} and FFHQ~\cite{karras2019style}. CelebA-HQ contains 30,000 aligned celebrity images at 1024×1024 resolution, offering clean visual statistics for controlled assessment of identity preservation and visual fidelity. FFHQ provides 70,000 images at the same resolution, with significantly greater diversity across age, ethnicity, pose, lighting, and background. This broad variability makes FFHQ a challenging testbed for evaluating the robustness and generalizability of facial privacy protection models. We randomly select 1,000 distinct-identity images as source images for both the FFHQ and CelebA-HQ datasets.

\textbf{Benchmark.} We evaluate the attack performance of SIDeR against a comprehensive set of adversarial baselines, including five noise-based approaches such as FGSM~\cite{goodfellow2014explaining}, MI-FGSM~\cite{dong2018boosting}, PGD~\cite{madry2017towards}, TI-DIM~\cite{dong2019evading}, TIP-IM~\cite{yang2021towards}, and twelve unrestricted approaches. The unrestricted methods consist of seven makeup-based techniques including Adv-Hat~\cite{komkov2021advhat}, Adv-Makeup~\cite{yin2021adv}, AMT-GAN~\cite{hu2022protecting}, Clip2Protect~\cite{shamshad2023clip2protect}, GIFT~\cite{li2024transferable}, DFPP~\cite{shamshad2024makeup}, and DiffAM~\cite{sun2024diffam}, and six face-semantics-invariant techniques including DiffProtect~\cite{liu2023diffprotect}, DPG~\cite{zhang2024double}, SD4Privacy~\cite{an2024sd4privacy}, Adv-Diffusion~\cite{liu2024adv}, P3-MASK~\cite{chow2025personalized}, and Adv-CPG~\cite{wang2025adv}. In addition, we compare SIDeR with AVIH~\cite{su2023hiding} and Diff-Privacy~\cite{he2024diff} to evaluate its capability in image information embedding and reversible reconstruction.

\begin{figure*}[htbp]
    \centering
    \includegraphics[width=1\linewidth]{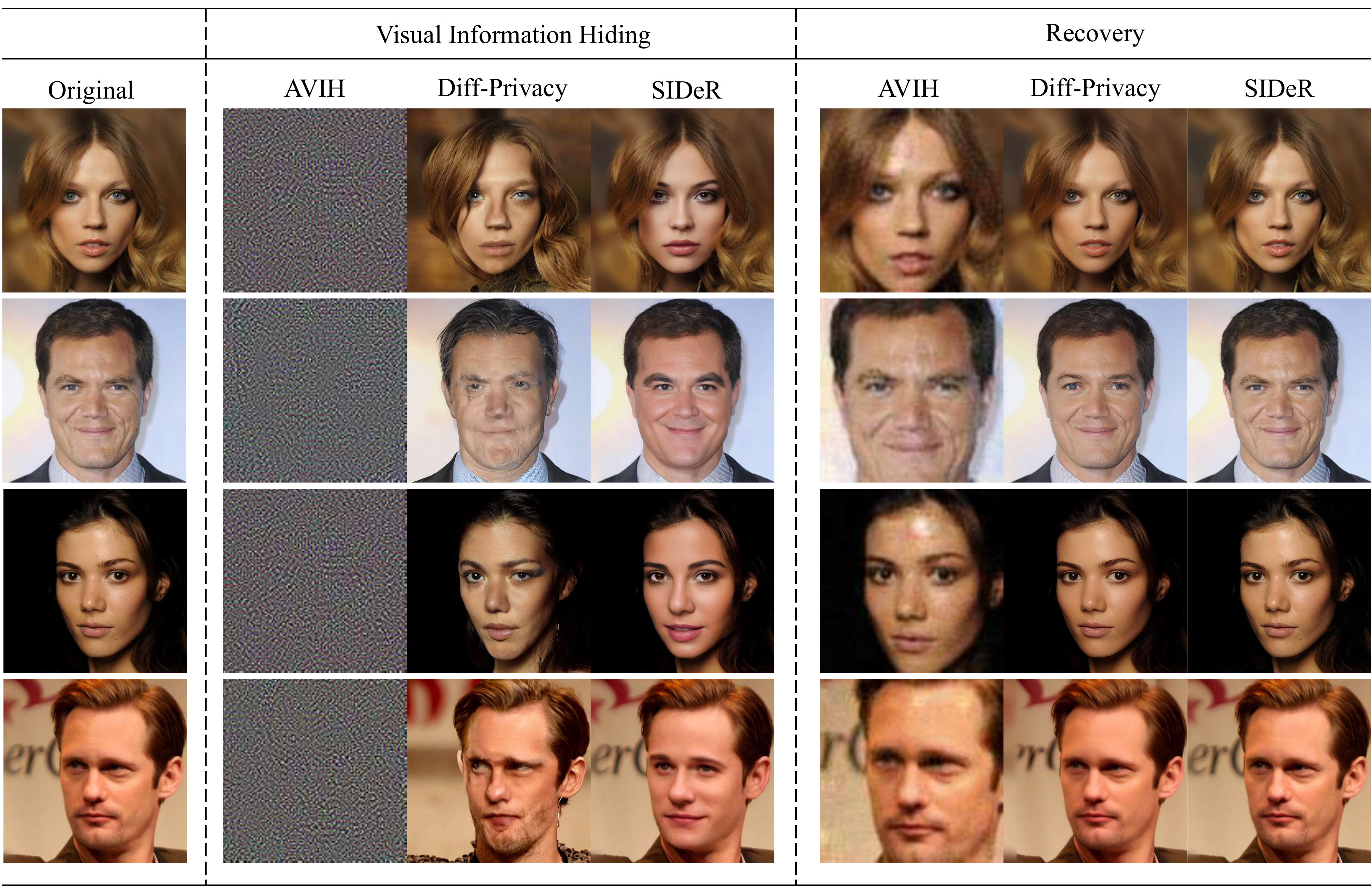}
    \caption{Visual comparison of image recovery quality. Compared to AVIH and Diff-Privacy, our proposed SIDeR significantly reduces blurring and artifacts, maintaining superior structural fidelity and visual details on both FFHQ and CelebA-HQ datasets.}
    \label{fig:recover_compare}
\end{figure*}

\textbf{Target Models.} Following Adv-CPG\cite{wang2025adv}, we assess the attack performance of SIDeR and state-of-the-art baselines under black-box conditions using four widely adopted deep face recognition models: Facenet~\cite{schroff2015facenet}, IR152~\cite{he2016deep}, IRSE50~\cite{hu2018squeeze}, and MobileFace~\cite{deng2019arcface}. In addition to these academic models, we further include two commercial face recognition services, Face++\footnote{https://www.faceplusplus.com/face-comparing} and Aliyun\footnote{\url{https://vision.aliyun.com/experience/detail?&tagName=facebody&children=CompareFace}}, to comprehensively evaluate the robustness of different adversarial attack methods in real-world deployment environments.

\textbf{Evaluation Metrics.} Our evaluation is from two complementary perspectives: privacy protection and image quality. For privacy protection, we assess face verification performance using the Attack Success Rate (ASR), defined as the proportion of adversarial samples that successfully mislead the recognition model. In commercial face recognition services, the confidence scores returned by the platforms are used directly to evaluate the effectiveness of different attack methods. To evaluate visual fidelity, we employ three standard metrics: FID, PSNR, and SSIM. Lower FID values and higher PSNR and SSIM values indicate greater visual similarity between the adversarial image and its reference.

\begin{table*}[!t]
    \caption{Attack Success Rate (ASR\%) of various adversarial methods in the black-box face verification setting. For each target model, adversarial examples are generated using the remaining three models as surrogates. Cells highlighted in \colorbox{yellow}{Dark yellow} indicate the highest performance, and those in \colorbox{yellow!40}{Light yellow} mark the second-highest results.}
    \centering
    \renewcommand{\arraystretch}{1.2}
    \scalebox{0.95}{\begin{tabular}{c|c|cccc|cccc|c}
        \toprule
        \textbf{Method} & \textbf{Dataset} & \multicolumn{4}{c}{\textbf{FFHQ}} & \multicolumn{4}{c|}{\textbf{CelebA-HQ}} &\textbf{Average}\\
         &ASR\% on FR Model $\uparrow$ & IR152  & IRSE50  & FaceNet & MobileFace & IR152  & IRSE50  & FaceNet & MobileFace   & \\
        \midrule
        Clean &— & 3.14 & 4.73 & 1.49 & 8.96 & 4.64 & 5.74 & 1.06 & 13.27 & 5.38\\
        \midrule
        &FGSM~\cite{goodfellow2014explaining} & 9.86 & 48.53 & 4.23 & 51.48 & 12.07 & 45.81 & 1.35 & 53.02 & 28.29\\
        &MI-FGSM~\cite{dong2018boosting} & 46.31 & 69.24 & 20.62 & 69.26 & 46.53 & 70.57 & 27.09 & 58.94 & 51.07\\
        Noise- &PGD~\cite{madry2017towards} & 31.64 & 75.38 & 18.59 & 63.12 & 41.87 & 63.23 & 19.62 & 57.34 & 46.35 \\
        Based&TI-DIM~\cite{dong2019evading} & 43.57 & 65.89 & 14.72 & 53.31 & 35.13 & 62.38 & 13.68 & 52.84 & 42.69\\
        &TIP-IM~\cite{yang2021towards} & 46.25 & 67.38 & 59.82 & 52.03 & 41.26 & 57.29 & 39.07 & 49.56 & 51.58\\
        \cmidrule{1-11}
        &Adv-Hat~\cite{komkov2021advhat} & 13.77 & 15.36 & 5.26 & 9.83 & 5.04 & 16.88 & 4.91 & 12.64 & 10.46\\
        &Adv-Makeup~\cite{yin2021adv} & 10.03 & 25.57 & 1.08 & 20.38 & 12.68 & 19.95 & 1.37 & 22.11 & 14.15\\
        &AMT-GAN~\cite{hu2022protecting} & 11.52 & 56.02 & 9.74 & 41.32 & 12.09 & 53.26 & 4.87 & 47.95 & 29.60\\
        Makeup&Clip2Protect~\cite{shamshad2023clip2protect} & 52.12 & 86.53 & 45.01 & 76.29 & 47.63 & 80.96 & 42.57 & 73.64 & 63.09\\
        -Based&GIFT~\cite{li2024transferable} & 69.72 & 87.64 & 54.49 & 82.93 & 73.84 & 83.72 & 56.48 & 86.37 & 74.40\\
        &DFPP~\cite{shamshad2024makeup} & 54.25 & 90.63 & 52.13 & 80.09 & 46.38 & 80.59 & 45.37 & 72.13 & 69.20\\
        &DiffAM~\cite{sun2024diffam} & 67.24 & 90.30 & \cellcolor{yellow!40}{64.96} & 89.56 & \cellcolor{yellow!40}{65.09} & \cellcolor{yellow!40}{89.66} & 62.99 & 84.51 & 76.79 \\
        \cmidrule{1-11}
        &DiffProtect~\cite{liu2023diffprotect} & 57.62 & 60.14 & 49.38 & 67.52 & 58.64 & 79.34 & 24.69 & 75.91 & 59.16\\
        Facial&DPG~\cite{zhang2024double} & 34.87 & 76.82 & 36.57 & 69.03 & 42.89 & 62.47 & 35.83 & 66.42 & 53.11\\
        Semantic&SD4Privacy~\cite{an2024sd4privacy} & 51.31 & 79.94 & 43.57 & 71.55 & 66.89 & 79.96 & 53.49 & 74.58 & 65.16\\
        Invariant&Adv-Diffusion~\cite{liu2024adv} & 50.93 & 81.76 & 30.84 & 67.52 & 52.84 & 81.67 & 34.95 & 70.78 & 58.91\\
        &P3-Mask~\cite{chow2025personalized} & 73.18 & 85.35 & 57.92 & 70.69 & 73.47 & 83.40 & 60.24 & 69.64 & 71.74\\
        &Adv-CPG~\cite{wang2025adv} & \cellcolor{yellow!40}{75.26} & \cellcolor{yellow!40}{91.03} & 63.84 & \cellcolor{yellow!40}{89.94} & \cellcolor{yellow!40}{76.96} & 88.72 & \cellcolor{yellow!40}{63.50} & \cellcolor{yellow!40}{87.95} & \cellcolor{yellow!40}{79.65}\\
        \cmidrule{1-11}
         &SIDeR(Cover)& \cellcolor{yellow}{92.70} & \cellcolor{yellow}{98.60} & \cellcolor{yellow}{96.30} & \cellcolor{yellow}{99.40} & \cellcolor{yellow}{96.30}  & \cellcolor{yellow}{99.00} & \cellcolor{yellow}{96.60} & \cellcolor{yellow}{99.70} & \cellcolor{yellow}{97.33}\\
        Semantic &SIDeR(Decoy)        & 93.10 & 98.30 &96.50 & 98.40 & 95.40 & 99.10 & 97.00 & 99.50 &97.16\\
        Decoupling &SIDeR(Hidden)     & 92.60 & 98.60 &96.50 & 99.50 & 96.30 & 99.10 & 96.60 & 99.70 &97.36\\
        &SIDeR(Recovery-Unauthorized) & 93.10 & 98.30 &96.70 & 98.60 & 95.50 & 99.20 & 97.20 & 99.50 &97.26\\

        \bottomrule
    \end{tabular}}
    \label{table:attack}
\end{table*}

\subsection{Comparison Study}
In this section, we conduct a systematic evaluation of the proposed method against a set of benchmark approaches under black-box settings across four pretrained face recognition models. Our assessment focuses on two key aspects: attack performance and the visual quality of both the generated and recovered images.

\textbf{Comparison on Black-Box Attacks.} Table \ref{table:attack} presents a comprehensive performance analysis of SIDeR and 18 competing methods on the FFHQ and CelebA-HQ benchmarks, encompassing both noise-based and unconstrained generative paradigms. For these face verification trials, the decision threshold $\tau$ is consistently determined by FAR@0.01, with the following model-specific settings: 0.167 for IR152, 0.241 for IRSE50, 0.409 for FaceNet, and 0.302 for MobileFace. In terms of average attack success rate, SIDeR demonstrates clear superiority. Compared with the strongest noise-based perturbation method, SIDeR achieves an absolute gain of approximately 28\%, and it further surpasses unconstrained generative methods by more than 3\%, indicating a stable advantage across different attack categories. In addition, because FaceNet's feature representations differ from those of ArcFace-based models, most existing methods exhibit poor transferability when attacking FaceNet. In contrast, SIDeR consistently maintains a high ASR on this architecture, revealing stronger cross-architecture transferability of the generated perturbations.

\begin{table*}[htbp]
  \caption{Quantitative comparison with advanced methods and detailed performance analysis of our proposed SIDeR on FFHQ and CelebA-HQ datasets. The best results for the recovery performance are highlighted."$\uparrow$" means bigger is better, and "$\downarrow$" means smaller is better. Cells highlighted in \colorbox{yellow}{Dark yellow} indicate the highest performance.}
  \label{table:overall_comparison}
  \centering
  \renewcommand{\arraystretch}{1.2}
  \scalebox{1.22}{
  \begin{tabular}{l|cccc|cccc}
    \toprule
    \multirow{2}{*}{\textbf{Method / Image Pairs}} & \multicolumn{4}{c|}{\textbf{FFHQ}} & \multicolumn{4}{c}{\textbf{CelebA-HQ}} \\
    & PSNR$\uparrow$ & SSIM$\uparrow$ & MSE$\downarrow$ & RMSE$\downarrow$ & PSNR$\uparrow$ & SSIM$\uparrow$ & MSE$\downarrow$ & RMSE$\downarrow$ \\
    \midrule
    \multicolumn{9}{l}{\textit{\textbf{Advanced Methods}}} \\
    \cmidrule{1-9}
    AVIH~\cite{su2023hiding} & 26.643 & 0.827 & 7.872 & 10.653 & 27.851 & 0.846 & 7.547 & 10.493 \\
    Diff-Privacy~\cite{he2024diff} & 27.747 & 0.766 & 7.135 & 11.769 & 28.238 & 0.789 & 6.791 & 10.257 \\
    \midrule
    \multicolumn{9}{l}{\textit{\textbf{Our Proposed SIDeR}}} \\
    \cmidrule{1-9}
    Cover / Hidden & 42.648 & 0.981 & 1.239 & 1.927 & 43.701 & 0.979 & 1.188 & 1.688 \\
    Decoy / Recovery-Unauthorized & 42.692 & 0.984 & 1.236 & 1.928 & 42.837 & 0.984 & 1.245 & 1.897 \\
    Secret / Recovery-Authorized & \cellcolor{yellow}{39.899} & \cellcolor{yellow}{0.973} & \cellcolor{yellow}{1.624} & \cellcolor{yellow}{2.675} & \cellcolor{yellow}{39.316} & \cellcolor{yellow}{0.967} & \cellcolor{yellow}{1.792} & \cellcolor{yellow}{2.915} \\
    \bottomrule
  \end{tabular}}
\end{table*}

\begin{table*}[htbp]
 \caption{Performance comparisons of our SIDeR method with nine other advanced adversarial attack and privacy protection strategies (including PGD, TIP-IM, Adv-Makeup, Adv-Diffusion, etc.) on FID $\downarrow$.} 
 \centering
 \renewcommand{\arraystretch}{1.2}
 \scalebox{1}{\begin{tabular}{c|cc|cccc|ccc|c}
 \toprule
 \textbf{Metric} & PGD & TIP-IM & Adv-Makeup & AMT-GAN & CLIP2Protect & DiffAM & SD4Privacy & Adv-Diffusion & Adv-CPG & SIDeR\\
 \cmidrule{1-11}
 FID$\downarrow$ & 78.92 & 38.7325 & 4.2282 & 34.5703  & 26.1272  & 26.1015 & 26.4078 & 22.5751  & 26.0758 & 24.7633\\
 \bottomrule
 \end{tabular}}
 \label{table:fid}
\end{table*}

\begin{figure}[htbp]
    \centering
    \includegraphics[width=1\linewidth]{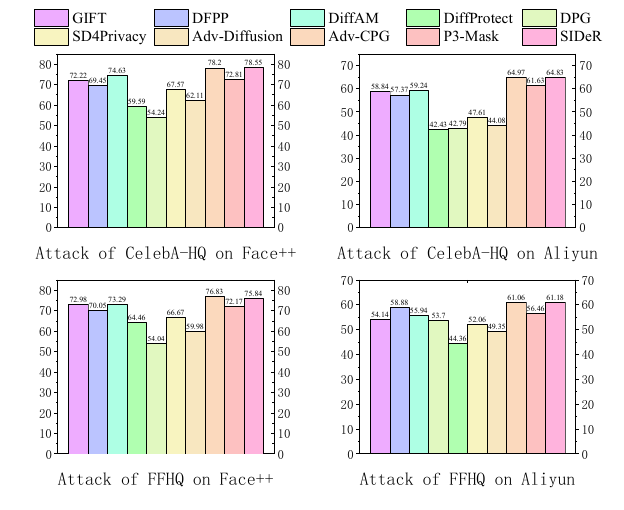}
    \caption{Confidence scores (↑) returned by two commercial face recognition APIs (Face++ and Aliyun) on adversarial examples generated by different privacy-preserving methods. SIDeR achieves confidence scores that are consistently comparable to or higher than those of state-of-the-art unrestricted makeup and semantic-invariant approaches, demonstrating strong effectiveness and stability across both datasets and APIs.}
    \label{fig:api}
\end{figure}

\textbf{Attack Performance on Commercial APIs.} Fig.~\ref {fig:api} summarizes the quantitative comparison results of SIDeR with nine state-of-the-art privacy-preserving techniques (covering two main categories: unrestricted makeup and semantic invariants) on commercial APIs. The evaluation metric uses the API's confidence score, which ranges from 0 to 100. A higher score indicates greater similarity between the adversarial example and the target image, i.e., a stronger privacy-preserving attack. The evaluation samples are 100 images randomly selected from the CelebA-HQ and FFHQ datasets. Overall, SIDeR demonstrates excellent stability and superior performance across all settings. 

In all four test scenarios, SIDeR maintains a top-tier performance, with its confidence score approaching or exceeding (i.e., the attack effect is close to or better than) the best-performing Adv-CPG. Notably, SIDeR exhibits excellent performance stability across different datasets and APIs, with a maximum score difference of only $0.99$ compared to Adv-CPG. This stable attack performance demonstrates SIDeR's strong black-box migration capabilities, indicating that its semantic-identity decoupling mechanism can effectively address privacy protection scenarios involving unknown architectures and real-world training data.

\textbf{Comparison on Image Quality.} 
We evaluate the perceptual fidelity of SIDeR against several state-of-the-art baselines, as visualized in Fig.~\ref{fig:attack}. Specifically, we compared SIDeR with two representative methods from the noise-based, unconstrained makeup, and semantic-invariant generative categories to highlight qualitative differences. In contrast, SIDeR better suppresses visual degradation from noise during generation and produces more natural facial texture and lighting. For makeup-based attack and defense methods, AMT-GAN and DiffAM explicitly incorporate perturbations into makeup styles. While such modifications are easy to evade traditional detection, they often alter the user's original appearance features, resulting in pronounced visual changes. Semantic-level attack methods (e.g., Adv-Diffusion and Adv-CPG) constrain facial changes by manipulating the latent space, thereby producing smoother overall edits. However, these methods are constrained by semantic preservation requirements, and typically allow only limited modifications to the original appearance, making it difficult to meet users' diverse needs for identity weakening, style preservation, or content customization. In contrast, SIDeR, through semantic-identity decoupling and local latent-space manipulation, weakens identity features while maintaining appearance consistency. Visually, SIDeR's output is closer to realistic images in facial details, background consistency, and overall naturalness, thus outperforming most comparison methods in subjective quality.

Table~\ref{table:fid} provides a quantitative analysis of image quality metrics on the CelebA-HQ dataset. Adv-Makeup achieves the lowest FID score because its modifications are minimal, only affecting the eyes; however, this method has a low ASR and cannot serve as an effective defense. In contrast, SIDeR achieves a relatively low FID score even without deliberately compressing the editing scope. This demonstrates that SIDeR can generate high-quality adversarial examples that are semantically consistent and exhibit natural structure while maintaining attack effectiveness.

From both visual and quantification perspectives, SIDeR achieves a more balanced performance across naturalness, fidelity, and controllability, maintaining excellent image quality while meeting attack requirements.

\textbf{Image Information Hiding and Recovery.} We evaluate the performance of SIDeR in image information hiding and multi-level recovery tasks. The experiments are conducted in two aspects: first, comparing the differences in end-to-end hiding and recovery capabilities between SIDeR, AVIH, and Diff-Privacy; second, analyzing the image quality at different recovery stages in the multi-layer structure of SIDeR, including the final hidden image, the decoy layer recovered image, and the secret image recovery results. Visual examples are shown in Fig.~\ref{fig:recover_compare}  and ~\ref{fig:recover_self}, and the corresponding quantitative indicators are listed in Table~\ref{table:overall_comparison}.

First, we observe the overall hiding quality. SIDeR maintains high image fidelity even after completing multi-layer embedding. Taking the FFHQ dataset as an example, the PSNR between the final hidden image and the original image reaches 42.648, and the SSIM is 0.981. On CelebA-HQ, the PSNR and SSIM reach 43.701 and 0.979, respectively. This demonstrates that, after embedding the secret image in the carrier image, SIDeR preserves the appearance of the original user-uploaded photo without causing significant visual degradation. In contrast, AVIH and Diff-Privacy only achieve PSNRs between 26 and 28 on the same task, and SSIM remains at 0.76 to 0.84. Visual results further reveal significant blurring and damage to content structure. Both quantitative metrics and visual results indicate that SIDeR significantly outperforms existing methods in  concealment quality.

\begin{figure}[htbp]
    \centering
    \includegraphics[width=1\linewidth]{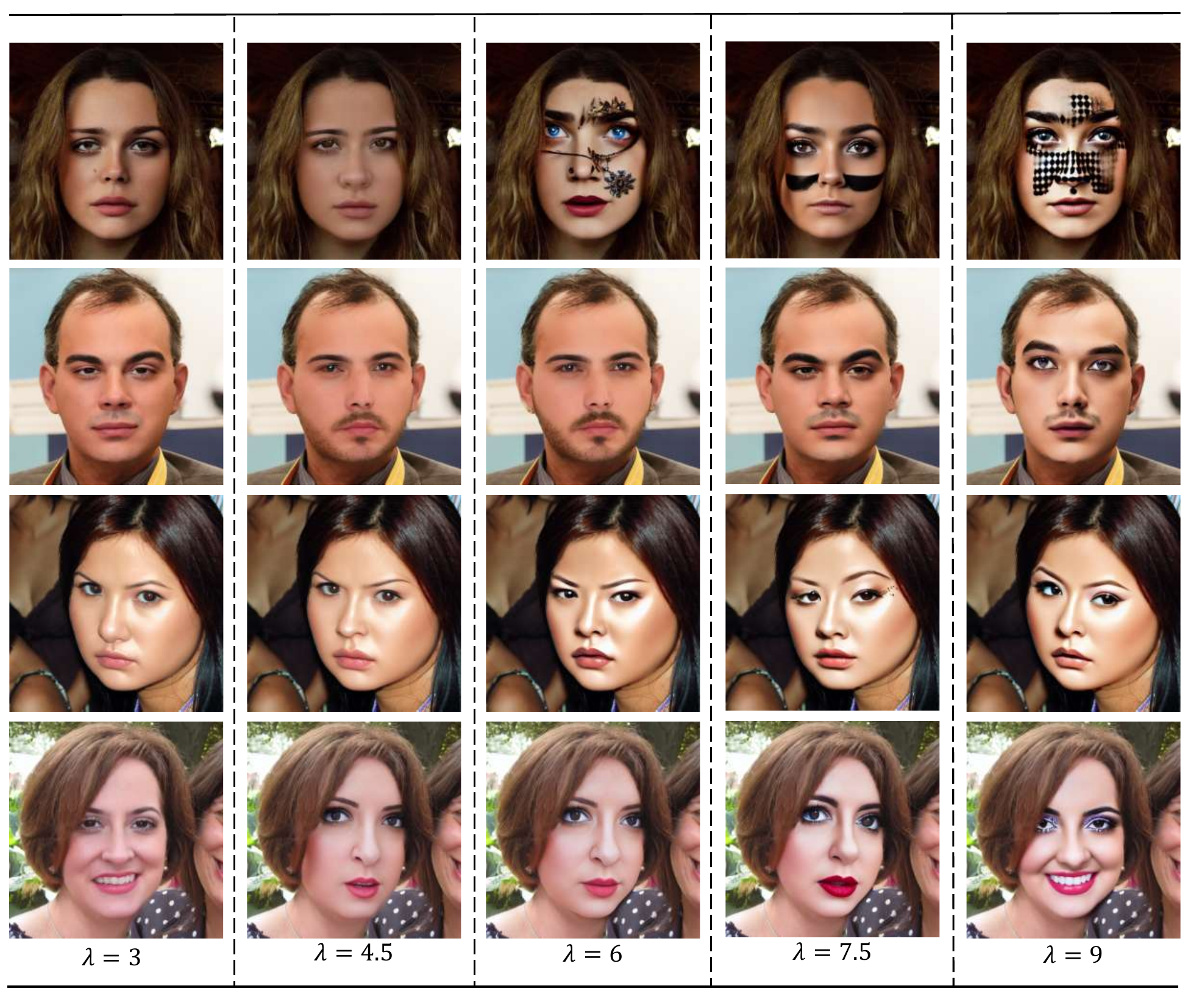}
    \caption{The resulting FFHQ images as $\lambda$ varies from $3$ to $9$. Notably, high values of $\lambda$ (e.g., $\lambda=9$) introduce conspicuous visual artifacts and geometric distortions to the facial features, confirming the essential role of $\lambda$ in regulating the perceptual naturalness of the protected image.}
    \label{fig:lambda_visual}
\end{figure}

Next, the decoy layer recovery effect is evaluated. SIDeR's structure allows for the reconstruction of a pre-designed decoy version before the complete recovery of the secret image. This serves as a safe intermediate form to mislead adversaries in potential decoding attacks. Experiments show that this intermediate result still maintains high image quality. On FFHQ, the PSNR and SSIM recovered by the decoy layer are 42.692 and 0.984, respectively; on CelebA-HQ, they are 42.837 and 0.984, respectively. Visual comparison also shows that the intermediate recovery results maintains stable consistency with the real image in overall structure, lighting, and facial details. This indicates that SIDeR's hierarchical design not only improves security but also ensures that the content at each stage of the recovery chain maintains reliable visual quality.

Finally, the recovery quality of the secret image is analyzed. SIDeR still achieved stable and reliable results in the final recovery stage. The PSNR on FFHQ is 39.899 and SSIM is 0.973; on CelebA-HQ, they are 39.316 and 0.967, respectively. Although the values at this stage are lower than those of the decoy layer, they are still substantially higher than the best levels achieved by existing methods, and the recovered secret image exhibits high consistency in identity features, lighting structure, and background texture. This demonstrates that multi-layer embedding does not cause cumulative damage and proves that SIDeR can reliably separate complete secret information from the final hidden image.

In summary, SIDeR achieves significant advantages in three dimensions: image quality, structural consistency, and multi-layer recovery capability. Irrespective of the recovery results for the carrier, decoy, or secret image, SIDeR substantially outperforms current baseline methods across all four metrics: PSNR, SSIM, MSE, and RMSE. Visualization results further confirm this conclusion: SIDeR can achieve high-quality information hiding and recovery while maintaining natural visual effects. This stability and accuracy make SIDeR more usable in real-world privacy scenarios.

\begin{figure}[htbp]
    \centering
    \includegraphics[width=1\linewidth]{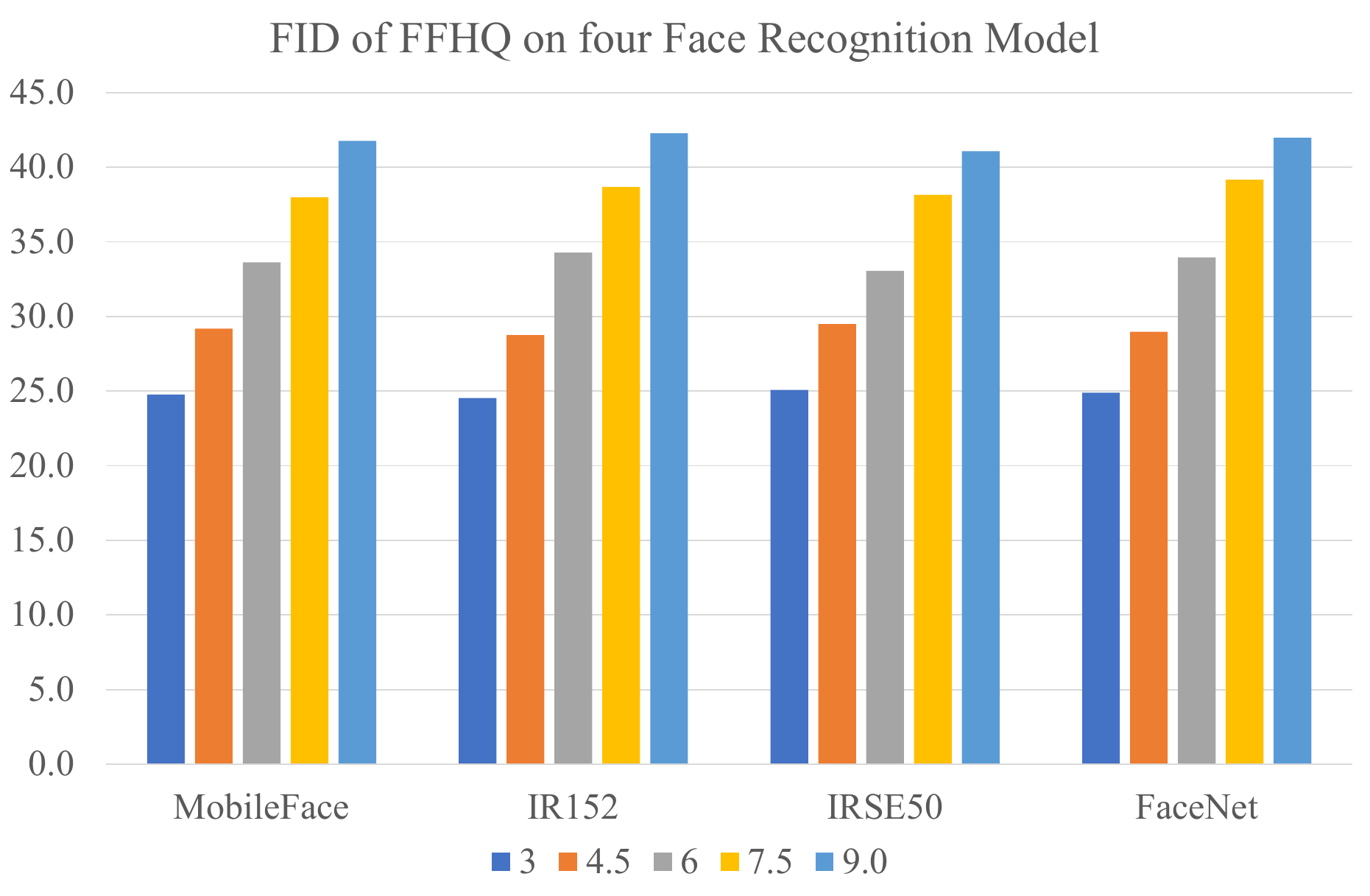}
    \caption{FID (Fréchet Inception Distance) metric under different values of $\lambda$. As $\lambda$ increases, the FID metric rises, indicating that while adversarial perturbation is enhanced, the visual fidelity of the image decreases.}
    \label{fig:lambda}
\end{figure}
\section{Ablation Study}
We also conduct experiments to analyze the contributions of $\lambda$ and the momentum factor. To evaluate the influence of the modulation coefficient $\lambda$ on the visual quality of the generated adversarial samples, we calculate the FID for images generated with different $\lambda$ values. As shown in Fig.~\ref{fig:lambda}, increasing the value of $\lambda$ generally leads to an increase in the FID score, indicating a reduction in image fidelity and perceptual quality. Fig. ~\ref{fig:lambda_visual}  provides a visual comparison of the faces generated under varying $\lambda$ settings. It is evident that high $\lambda$ values often introduce noticeable artifacts, unnatural geometric distortions, or unusual facial features. This observation confirms the critical role of $\lambda$ in balancing the adversarial perturbation strength against the visual naturalness of the protected image. The findings suggest that precise tuning of $\lambda$ is essential to achieve high attack efficacy while maintaining satisfactory visual quality.

\begin{table}[htbp]
  \caption{We compared the attack success rates (ASR $\uparrow$) of the full SIDeR model and the variant with the momentum term removed ('w/o momentum') on the FFHQ dataset. The results show that the introduction of the momentum term improves the attack's strength and stability. Cells highlighted in \colorbox{yellow}{Dark yellow} indicate the highest performance.}
  \label{table:mu}
  \centering
  \renewcommand{\arraystretch}{1.2}
  \scalebox{1}{
  \begin{tabular}{l|cccc}
    \toprule
    \textbf{Method(ASR$\uparrow$)}  & IR152  & IRSE50  & FaceNet & MobileFace \\
    \midrule
    \cmidrule{1-5}
    SIDeR(w/o  momentum) & 92.10 & 95.60 & 95.16 & 97.50 \\
    SIDeR                & \cellcolor{yellow}{92.70} & \cellcolor{yellow}{98.60} & \cellcolor{yellow}{96.30} & \cellcolor{yellow}{99.40}\\
    \bottomrule
  \end{tabular}}
\end{table}

We further conduct an ablation study by setting the momentum factor to 0 (i.e., $\mu=0$) to assess its contribution to adversarial optimization. Table~\ref{table:mu} compares the ASR of the full SIDeR model with the variant without the momentum term (denoted as `SIDeR w/o momentum') on the FFHQ and CelebA-HQ datasets. The results clearly indicate a degradation in ASR across all evaluated target models when the momentum factor is removed. This suggests that the momentum term plays a crucial role by accumulating gradient history and stabilizing the update direction, thereby mitigating oscillations and enabling the optimization process to converge to superior adversarial solutions rapidly and robustly. Thus, the incorporation of the momentum factor $\mu$ is critical for maximizing attack strength and transferability.

\section{Conclusion}
In this paper, we have designed a novel reversible and semantically controllable framework, SIDeR, to achieve robust facial privacy protection while maintaining the utility of machine-level identity verification. SIDeR benefits from SD-AGM and CRM. SD-AGM leverages momentum-based latent-space optimization within a diffusion model to generate visually distinct adversarial covers and decoys that preserve the original identity features for recognition models. CRM embeds the original facial information into these carriers using a nested invertible neural network, ensuring that the original image can be near-losslessly recovered when the correct password is provided. When an incorrect password is provided, the module executes a deceptive recovery process. This process produces a realistic adversarial decoy to fool unauthorized users into believing the decryption was successful, thereby shielding the actual identity from exposure. Extensive experiments on FFHQ and CelebA-HQ have revealed that our scheme outperforms existing baselines and demonstrates strong performance in visual anonymity, identity preservation, and authorized high-fidelity reconstruction of original images. As a potential future direction, we are looking forward to extending our method to improve the performance of various applications such as large language models~\cite{fang2024automated,lin2023pushing} and distributed learning system~\cite{lin2025leo,zhang2024satfed,wei2025optimizing,lyu2023optimal,lin2025hasfl,zhang2025satfed}.

%

\bibliographystyle{IEEEtran}
\bibliography{IEEEabrv, ref}

\begin{thebibliography}{10}
\providecommand{\url}[1]{#1}
\csname url@samestyle\endcsname
\providecommand{\newblock}{\relax}
\providecommand{\bibinfo}[2]{#2}
\providecommand{\BIBentrySTDinterwordspacing}{\spaceskip=0pt\relax}
\providecommand{\BIBentryALTinterwordstretchfactor}{4}
\providecommand{\BIBentryALTinterwordspacing}{\spaceskip=\fontdimen2\font plus
\BIBentryALTinterwordstretchfactor\fontdimen3\font minus \fontdimen4\font\relax}
\providecommand{\BIBforeignlanguage}[2]{{%
\expandafter\ifx\csname l@#1\endcsname\relax
\typeout{** WARNING: IEEEtran.bst: No hyphenation pattern has been}%
\typeout{** loaded for the language `#1'. Using the pattern for}%
\typeout{** the default language instead.}%
\else
\language=\csname l@#1\endcsname
\fi
#2}}
\providecommand{\BIBdecl}{\relax}
\BIBdecl

\bibitem{sun2025rrto}
Z.~Sun, X.~Guan, Z.~Lin, Y.~Qing, H.~Song, Z.~Fang, Z.~Chen, F.~Liu, H.~Cui, W.~Ni \emph{et~al.}, ``Rrto: A high-performance transparent offloading system for model inference in mobile edge computing,'' \emph{arXiv preprint arXiv:2507.21739}, 2025.

\bibitem{duan2025leed}
T.~Duan, Z.~Zhang, S.~Guo, D.~Huang, Y.~Zhao, Z.~Lin, Z.~Fang, D.~Luan, H.~Cui, and Y.~Cui, ``Leed: A highly efficient and scalable llm-empowered expert demonstrations framework for multi-agent reinforcement learning,'' \emph{arXiv preprint arXiv:2509.14680}, 2025.

\bibitem{yuan2025constructing}
H.~Yuan, Z.~Chen, Z.~Lin, J.~Peng, Y.~Zhong, X.~Hu, S.~Xue, W.~Li, and Y.~Gao, ``{Constructing 4D Radio Map in LEO Satellite Networks with Limited Samples},'' \emph{{IEEE} INFOCOM}, 2025.

\bibitem{zhao2024leo}
Z.~Zhao, Z.~Chen, Z.~Lin, W.~Zhu, K.~Qiu, C.~You, and Y.~Gao, ``{LEO Satellite Networks Assisted Geo-Distributed Data Processing},'' \emph{{IEEE} Wireless Commun. Lett.}, 2024.

\bibitem{lin2025sl}
Z.~Lin, Z.~Lin, M.~Yang, J.~Huang, Y.~Zhang, Z.~Fang, X.~Du, Z.~Chen, S.~Zhu, and W.~Ni, ``Sl-acc: A communication-efficient split learning framework with adaptive channel-wise compression,'' \emph{arXiv preprint arXiv:2508.12984}, 2025.

\bibitem{fang2025dynamic}
Z.~Fang, Z.~Lin, S.~Hu, Y.~Tao, Y.~Deng, X.~Chen, and Y.~Fang, ``Dynamic uncertainty-aware multimodal fusion for outdoor health monitoring,'' \emph{arXiv preprint arXiv:2508.09085}, 2025.

\bibitem{zhang2025robust}
Z.~Zhang, T.~Duan, Z.~Lin, D.~Huang, Z.~Fang, Z.~Sun, L.~Xiong, H.~Liang, H.~Cui, Y.~Cui \emph{et~al.}, ``Robust deep reinforcement learning in robotics via adaptive gradient-masked adversarial attacks,'' \emph{arXiv preprint arXiv:2503.20844}, 2025.

\bibitem{NIPS2014_f033ed80}
\BIBentryALTinterwordspacing
I.~J. Goodfellow, J.~Pouget-Abadie, M.~Mirza, B.~Xu, D.~Warde-Farley, S.~Ozair, A.~Courville, and Y.~Bengio, ``Generative adversarial nets,'' in \emph{Advances in Neural Information Processing Systems}, Z.~Ghahramani, M.~Welling, C.~Cortes, N.~Lawrence, and K.~Weinberger, Eds., vol.~27.\hskip 1em plus 0.5em minus 0.4em\relax Curran Associates, Inc., 2014. [Online]. Available: \url{https://proceedings.neurips.cc/paper_files/paper/2014/file/f033ed80deb0234979a61f95710dbe25-Paper.pdf}
\BIBentrySTDinterwordspacing

\bibitem{lin2022channel}
Z.~Lin, L.~Wang, J.~Ding, B.~Tan, and S.~Jin, ``{Channel Power Gain Estimation for Terahertz Vehicle-to-Infrastructure Networks},'' \emph{{IEEE} Commun. Lett.}, vol.~27, no.~1, pp. 155--159, 2022.

\bibitem{NEURIPS2020_4c5bcfec}
\BIBentryALTinterwordspacing
J.~Ho, A.~Jain, and P.~Abbeel, ``Denoising diffusion probabilistic models,'' in \emph{Advances in Neural Information Processing Systems}, H.~Larochelle, M.~Ranzato, R.~Hadsell, M.~Balcan, and H.~Lin, Eds., vol.~33.\hskip 1em plus 0.5em minus 0.4em\relax Curran Associates, Inc., 2020, pp. 6840--6851. [Online]. Available: \url{https://proceedings.neurips.cc/paper_files/paper/2020/file/4c5bcfec8584af0d967f1ab10179ca4b-Paper.pdf}
\BIBentrySTDinterwordspacing

\bibitem{11245609}
S.~Wang, S.~Chen, D.-H. Wang, Y.~Hua, and Y.~Yan, ``Camd: Context-aware masked distillation for general self-supervised facial representation pre-training,'' \emph{IEEE Transactions on Circuits and Systems for Video Technology}, pp. 1--1, 2025.

\bibitem{gross2006model}
R.~Gross, L.~Sweeney, F.~De~la Torre, and S.~Baker, ``Model-based face de-identification,'' in \emph{2006 Conference on computer vision and pattern recognition workshop (CVPRW'06)}.\hskip 1em plus 0.5em minus 0.4em\relax IEEE, 2006, pp. 161--161.

\bibitem{meden2017kappa}
B.~Meden, Z.~Emersic, V.~Struc, and P.~Peer, ``$\kappa$-same-net: neural-network-based face deidentification,'' in \emph{2017 International Conference and Workshop on Bioinspired Intelligence (IWOBI)}.\hskip 1em plus 0.5em minus 0.4em\relax IEEE, 2017, pp. 1--7.

\bibitem{chen2007tools}
D.~Chen, Y.~Chang, R.~Yan, and J.~Yang, ``Tools for protecting the privacy of specific individuals in video,'' \emph{EURASIP Journal on Advances in Signal Processing}, vol. 2007, pp. 1--9, 2007.

\bibitem{butler2015privacy}
D.~J. Butler, J.~Huang, F.~Roesner, and M.~Cakmak, ``The privacy-utility tradeoff for remotely teleoperated robots,'' in \emph{Proceedings of the tenth annual ACM/IEEE international conference on human-robot interaction}, 2015, pp. 27--34.

\bibitem{gu2020password}
X.~Gu, W.~Luo, M.~S. Ryoo, and Y.~J. Lee, ``Password-conditioned anonymization and deanonymization with face identity transformers,'' in \emph{European conference on computer vision}.\hskip 1em plus 0.5em minus 0.4em\relax Springer, 2020, pp. 727--743.

\bibitem{li2023riddle}
D.~Li, W.~Wang, K.~Zhao, J.~Dong, and T.~Tan, ``Riddle: Reversible and diversified de-identification with latent encryptor,'' \emph{arXiv preprint arXiv:2303.05171}, 2023.

\bibitem{hukkelaas2019deepprivacy}
H.~Hukkel{\aa}s, R.~Mester, and F.~Lindseth, ``Deepprivacy: A generative adversarial network for face anonymization,'' in \emph{International symposium on visual computing}.\hskip 1em plus 0.5em minus 0.4em\relax Springer, 2019, pp. 565--578.

\bibitem{maximov2020ciagan}
M.~Maximov, I.~Elezi, and L.~Leal-Taix{\'e}, ``Ciagan: Conditional identity anonymization generative adversarial networks,'' in \emph{Proceedings of the IEEE/CVF conference on computer vision and pattern recognition}, 2020, pp. 5447--5456.

\bibitem{hu2022protecting}
S.~Hu, X.~Liu, Y.~Zhang, M.~Li, L.~Y. Zhang, H.~Jin, and L.~Wu, ``Protecting facial privacy: Generating adversarial identity masks via style-robust makeup transfer,'' in \emph{CVPR}, 2022, pp. 15\,014--15\,023.

\bibitem{shamshad2023clip2protect}
F.~Shamshad, M.~Naseer, and K.~Nandakumar, ``Clip2protect: Protecting facial privacy using text-guided makeup via adversarial latent search,'' in \emph{CVPR}, 2023, pp. 20\,595--20\,605.

\bibitem{li2024transferable}
M.~Li, J.~Wang, H.~Zhang, Z.~Zhou, S.~Hu, and pei Xiaobing, ``Transferable adversarial facial images for privacy protection,'' in \emph{ACM MM}, 2024.

\bibitem{shamshad2024makeup}
F.~Shamshad, M.~Naseer, and K.~Nandakumar, ``Makeup-guided facial privacy protection via untrained neural network priors,'' \emph{arXiv preprint arXiv:2408.12387}, 2024.

\bibitem{sun2024diffam}
Y.~Sun, L.~Yu, H.~Xie, J.~Li, and Y.~Zhang, ``Diffam: Diffusion-based adversarial makeup transfer for facial privacy protection,'' in \emph{CVPR}, 2024, pp. 24\,584--24\,594.

\bibitem{guan2022deepmih}
Z.~Guan, J.~Jing, X.~Deng, M.~Xu, L.~Jiang, Z.~Zhang, and Y.~Li, ``Deepmih: Deep invertible network for multiple image hiding,'' \emph{IEEE Transactions on Pattern Analysis and Machine Intelligence}, vol.~45, no.~1, pp. 372--390, 2022.

\bibitem{goodfellow2015explainingharnessingadversarialexamples}
\BIBentryALTinterwordspacing
I.~J. Goodfellow, J.~Shlens, and C.~Szegedy, ``Explaining and harnessing adversarial examples,'' 2015. [Online]. Available: \url{https://arxiv.org/abs/1412.6572}
\BIBentrySTDinterwordspacing

\bibitem{11267085}
X.~Du, X.~Liu, J.~Zhou, Z.~Lin, C.-m. Pun, C.~Wu, T.~Li, Z.~Chen, W.~Ni, and J.~Luo, ``Defensive adversarial captcha: A semantics-driven framework for natural adversarial example generation,'' \emph{IEEE Transactions on Dependable and Secure Computing}, pp. 1--13, 2025.

\bibitem{zhu2024dp}
J.~Zhu, X.~Du, J.~Zhou, C.-M. Pun, Q.~Xu, and X.~Liu, ``Dp-rae: A dual-phase merging reversible adversarial example for image privacy protection,'' in \emph{Proceedings of the 32nd ACM International Conference on Multimedia}, 2024, pp. 671--680.

\bibitem{goodfellow2014explaining}
I.~J. Goodfellow, J.~Shlens, and C.~Szegedy, ``Explaining and harnessing adversarial examples,'' in \emph{ICLR}, 2015.

\bibitem{dong2018boosting}
Y.~Dong, F.~Liao, T.~Pang, H.~Su, J.~Zhu, X.~Hu, and J.~Li, ``Boosting adversarial attacks with momentum,'' in \emph{CVPR}, 2018, pp. 9185--9193.

\bibitem{madry2017towards}
A.~Madry, A.~Makelov, L.~Schmidt, D.~Tsipras, and A.~Vladu, ``Towards deep learning models resistant to adversarial attacks,'' in \emph{ICLR}, 2018.

\bibitem{dong2019evading}
Y.~Dong, T.~Pang, H.~Su, and J.~Zhu, ``Evading defenses to transferable adversarial examples by translation-invariant attacks,'' in \emph{CVPR}, 2019, pp. 4312--4321.

\bibitem{yang2021towards}
X.~Yang, Y.~Dong, T.~Pang, H.~Su, J.~Zhu, Y.~Chen, and H.~Xue, ``Towards face encryption by generating adversarial identity masks,'' in \emph{ICCV}, 2021, pp. 3897--3907.

\bibitem{komkov2021advhat}
S.~Komkov and A.~Petiushko, ``Advhat: Real-world adversarial attack on arcface face id system,'' in \emph{ICPR}, 2021, pp. 819--826.

\bibitem{yin2021adv}
B.~Yin, W.~Wang, T.~Yao, J.~Guo, Z.~Kong, S.~Ding, J.~Li, and C.~Liu, ``Adv-makeup: A new imperceptible and transferable attack on face recognition,'' in \emph{IJCAI}, 2021, pp. 1252--1258.

\bibitem{liu2023diffprotect}
J.~Liu, C.~P. Lau, and R.~Chellappa, ``Diffprotect: Generate adversarial examples with diffusion models for facial privacy protection,'' \emph{arXiv preprint arXiv:2305.13625}, 2023.

\bibitem{zhang2024double}
Y.~Zhang, D.~Ye, S.~Shen, C.~Xie, Z.~Liu, J.~Deng, and L.~Tang, ``Double privacy guard: Robust traceable adversarial watermarking against face recognition,'' \emph{arXiv preprint arXiv:2404.14693}, 2024.

\bibitem{an2024sd4privacy}
J.~An, W.~Zhang, D.~Wu, Z.~Lin, J.~Gu, and W.~Wang, ``Sd4privacy: Exploiting stable diffusion for protecting facial privacy,'' in \emph{ICME}, 2024, pp. 1--6.

\bibitem{liu2024adv}
D.~Liu, X.~Wang, C.~Peng, N.~Wang, R.~Hu, and X.~Gao, ``Adv-diffusion: imperceptible adversarial face identity attack via latent diffusion model,'' in \emph{AAAI}, vol.~38, no.~4, 2024, pp. 3585--3593.

\bibitem{chow2025personalized}
K.-H. Chow, S.~Hu, T.~Huang, and L.~Liu, ``Personalized privacy protection mask against unauthorized facial recognition,'' in \emph{ECCV}, 2025, pp. 434--450.

\bibitem{wang2025adv}
J.~Wang, H.~Zhang, and Y.~Yuan, ``Adv-cpg: A customized portrait generation framework with facial adversarial attacks,'' in \emph{Proceedings of the Computer Vision and Pattern Recognition Conference}, 2025, pp. 21\,001--21\,010.

\bibitem{su2023hiding}
Z.~Su, D.~Zhou, N.~Wang, D.~Liu, Z.~Wang, and X.~Gao, ``Hiding visual information via obfuscating adversarial perturbations,'' in \emph{Proceedings of the IEEE/CVF International Conference on Computer Vision}, 2023, pp. 4356--4366.

\bibitem{Wang_Liu_Luo_Yang_Wang_2022}
\BIBentryALTinterwordspacing
Y.~Wang, J.~Liu, M.~Luo, L.~Yang, and L.~Wang, ``Privacy-preserving face recognition in the frequency domain,'' \emph{Proceedings of the AAAI Conference on Artificial Intelligence}, vol.~36, no.~3, pp. 2558--2566, Jun. 2022. [Online]. Available: \url{https://ojs.aaai.org/index.php/AAAI/article/view/20157}
\BIBentrySTDinterwordspacing

\bibitem{10121472}
Y.~Zhang, T.~Wang, R.~Zhao, W.~Wen, and Y.~Zhu, ``Rapp: Reversible privacy preservation for various face attributes,'' \emph{IEEE Transactions on Information Forensics and Security}, vol.~18, pp. 3074--3087, 2023.

\bibitem{he2024diff}
X.~He, M.~Zhu, D.~Chen, N.~Wang, and X.~Gao, ``Diff-privacy: Diffusion-based face privacy protection,'' \emph{IEEE Transactions on Circuits and Systems for Video Technology}, 2024.

\bibitem{rombach2022high}
R.~Rombach, A.~Blattmann, D.~Lorenz, P.~Esser, and B.~Ommer, ``High-resolution image synthesis with latent diffusion models,'' in \emph{Proceedings of the IEEE/CVF conference on computer vision and pattern recognition}, 2022, pp. 10\,684--10\,695.

\bibitem{kingma2013auto}
D.~P. Kingma, M.~Welling \emph{et~al.}, ``Auto-encoding variational bayes,'' 2013.

\bibitem{li2023blip}
J.~Li, D.~Li, S.~Savarese, and S.~Hoi, ``Blip-2: Bootstrapping language-image pre-training with frozen image encoders and large language models,'' in \emph{International conference on machine learning}.\hskip 1em plus 0.5em minus 0.4em\relax PMLR, 2023, pp. 19\,730--19\,742.

\bibitem{achiam2023gpt}
J.~Achiam, S.~Adler, S.~Agarwal, L.~Ahmad, I.~Akkaya, F.~L. Aleman, D.~Almeida, J.~Altenschmidt, S.~Altman, S.~Anadkat \emph{et~al.}, ``Gpt-4 technical report (2023),'' \emph{arXiv preprint arXiv:2303.08774}, 2023.

\bibitem{CelebAMask-HQ}
C.-H. Lee, Z.~Liu, L.~Wu, and P.~Luo, ``Maskgan: Towards diverse and interactive facial image manipulation,'' in \emph{IEEE Conference on Computer Vision and Pattern Recognition (CVPR)}, 2020.

\bibitem{karras2019style}
T.~Karras, S.~Laine, and T.~Aila, ``A style-based generator architecture for generative adversarial networks,'' in \emph{Proceedings of the IEEE/CVF conference on computer vision and pattern recognition}, 2019, pp. 4401--4410.

\bibitem{schroff2015facenet}
F.~Schroff, D.~Kalenichenko, and J.~Philbin, ``Facenet: A unified embedding for face recognition and clustering,'' in \emph{Proceedings of the IEEE conference on computer vision and pattern recognition}, 2015, pp. 815--823.

\bibitem{he2016deep}
K.~He, X.~Zhang, S.~Ren, and J.~Sun, ``Deep residual learning for image recognition,'' in \emph{Proceedings of the IEEE conference on computer vision and pattern recognition}, 2016, pp. 770--778.

\bibitem{hu2018squeeze}
J.~Hu, L.~Shen, and G.~Sun, ``Squeeze-and-excitation networks,'' in \emph{Proceedings of the IEEE conference on computer vision and pattern recognition}, 2018, pp. 7132--7141.

\bibitem{deng2019arcface}
J.~Deng, J.~Guo, N.~Xue, and S.~Zafeiriou, ``Arcface: Additive angular margin loss for deep face recognition,'' in \emph{Proceedings of the IEEE/CVF conference on computer vision and pattern recognition}, 2019, pp. 4690--4699.

\bibitem{fang2024automated}
Z.~Fang, Z.~Lin, Z.~Chen, X.~Chen, Y.~Gao, and Y.~Fang, ``{Automated Federated Pipeline for Parameter-Efficient Fine-Tuning of Large Language Models},'' \emph{{IEEE} Trans. Mobile Comput.}, 2026.

\bibitem{lin2023pushing}
Z.~Lin, G.~Qu, Q.~Chen, X.~Chen, Z.~Chen, and K.~Huang, ``{Pushing Large Language Models to the 6G Edge: Vision, Challenges, and Opportunities},'' \emph{IEEE Communication Magazine}, 2023.

\bibitem{lin2025leo}
Z.~Lin, Y.~Zhang, Z.~Chen, Z.~Fang, C.~Wu, X.~Chen, Y.~Gao, and J.~Luo, ``{LEO-Split: A Semi-Supervised Split Learning Framework over LEO Satellite Networks},'' \emph{{IEEE} Trans. Mobile Comput.}, 2025.

\bibitem{zhang2024satfed}
Y.~Zhang, Z.~Lin, Z.~Chen, Z.~Fang, W.~Zhu, X.~Chen, J.~Zhao, and Y.~Gao, ``Satfed: A resource-efficient leo satellite-assisted heterogeneous federated learning framework,'' \emph{Engineering}, 2024.

\bibitem{wei2025optimizing}
W.~Wei, Z.~Lin, X.~Liu, H.~Du, D.~Niyato, and X.~Chen, ``Optimizing split federated learning with unstable client participation,'' \emph{arXiv preprint arXiv:2509.17398}, 2025.

\bibitem{lyu2023optimal}
S.~Lyu, Z.~Lin, G.~Qu, X.~Chen, X.~Huang, and P.~Li, ``Optimal resource allocation for u-shaped parallel split learning,'' in \emph{2023 IEEE Globecom Workshops (GC Wkshps)}, 2023, pp. 197--202.

\bibitem{lin2025hasfl}
Z.~Lin, Z.~Chen, X.~Chen, W.~Ni, and Y.~Gao, ``{HASFL: Heterogeneity-aware Split Federated Learning over Edge Computing Systems},'' \emph{arXiv preprint arXiv:2506.08426}, 2025.

\end{thebibliography}

\vfill

\end{document}